%% file: iclr2025_conference.tex
\documentclass{article} % For LaTeX2e
\usepackage{iclr2025_conference,times}
\usepackage{graphicx}
\usepackage{booktabs}
\usepackage{amsmath}
\usepackage{amsfonts}
\usepackage{amssymb}
\usepackage{multirow}
\usepackage{rotating}
\input{math_commands.tex}

\usepackage{hyperref}
\usepackage{url}

\title{Neuromorphic Principles for Efficient Large Language Models on Intel Loihi 2}

\author{
Steven Abreu\\
University of Groningen \& Intel Labs\\
\texttt{s.abreu@rug.nl} \\
\And
Sumit Bam Shrestha\\
Intel Labs\\
\texttt{sumit.bam.shrestha@intel.com} \\
\And
Rui-Jie Zhu \\
University of California, Santa Cruz \\
\texttt{rzhu48@ucsc.edu}\\
\And
Jason Eshraghian \\
University of California, Santa Cruz \\
\texttt{jsn@ucsc.edu}
}

\iclrfinalcopy % Uncomment for camera-ready version, but NOT for submission.
\begin{document}

\maketitle

\begin{abstract}
Large language models (LLMs) deliver impressive performance but require large amounts of energy. In this work, we present a MatMul‐free LLM architecture adapted for Intel's neuromorphic processor, Loihi 2. Our approach leverages Loihi 2's support for low-precision, event-driven computation and stateful processing. Our hardware-aware quantized model on GPU demonstrates that a 370M-parameter MatMul-free model can be quantized with no accuracy loss. Based on preliminary results, we report up to 3$\times$ higher throughput with 2$\times$ less energy, compared to transformer-based LLMs on an edge GPU, with significantly better scaling. Further hardware optimizations will increase throughput and decrease energy consumption. These results show the potential of neuromorphic hardware for efficient inference and pave the way for efficient reasoning models capable of generating complex, long-form text rapidly and cost-effectively.
\end{abstract}

\section{Introduction}

Large language models (LLMs) have revolutionized machine learning—but their computational and energy demands are enormous. This challenge motivates the development of efficient and scalable foundation models that are optimized not only algorithmically but co-designed with novel hardware architectures. 
In this paper, we propose a hardware-aware approach that integrates an efficient LLM architecture with Intel's neuromorphic processor, Loihi 2 \citep{davies_advancing_2021}. Although originally designed for event-based, sparse computations in spiking neural networks, Loihi 2's support for low-precision arithmetic and unstructured weight sparsity makes it an attractive platform for reducing energy consumption and latency in LLM inference. See Appendix \ref{app:l2-hw-arch} for more details on Loihi 2.

Although LLMs have been dominated by self-attention \citep{vaswani_attention_2017} with quadratic runtime complexity, LLMs based on state space models (SSMs) offer linear scaling with competitive performance \citep{gu2023mamba}. SSMs rely on element-wise recurrence \citep{gupta2022diagonal}, and use stateful neurons that align well with compute-near-memory architectures like Loihi 2.
Advances in quantization of LLMs \citep{dettmers_llmint8_2022,frantar_gptq_2023,xiao_smoothquant_2024} have culminated in extreme quantization at scale with LLMs using binary activations \citep{zhu2023spikegpt} or ternary weight matrices as seen in BitNet \citep{ma_era_2024}, piecewise affine transformers \citep{kosson_multiplication-free_2023}, ShiftAddLLM \citep{you_shiftaddllm_2024}, and earlier work on binarized neural networks \citep{courbariaux_binarized_2016}. 
Building on these ideas, \cite{zhu_scalable_2024} introduced the MatMul-free LLM, which replaces traditional matrix multiplications (MatMuls) with ternary matrices and element-wise operations, while also using a subquadratic SSM layer based on the HGRN model \citep{qin_hierarchically_2023,qin_hgrn2_2024}.

Loihi 2 is optimized for sequence-based processing, element-wise recurrence, low-precision arithmetic, and weight sparsity, all of which are features of the MatMul-free model. These benefits have been demonstrated on signal processing tasks \citep{orchard_efficient_2021,shrestha_efficient_2024} with orders of magnitude better latency and efficiency than state-of-the-art solutions.
This paper presents a work-in-progress of adapting and deploying the MatMul-free language model from \cite{zhu_scalable_2024} to Intel Loihi 2, opening a pathway that bridges neuromorphic computing with state-of-the-art efficient LLMs. 
This required a microcode implementation to map the MMF model to Loihi 2's architecture, along with a detailed ablation study to evaluate the optimal bit-precision for all operators in the MMF model. We are able to run the MMF model fully on-chip, using fixed-point arithmetic to optimize for energy and latency. 
% This work advances the recent effort to accelerate language models on neuromorphic systems, such as \cite{nazeer_language_2024}, who achieved 81.4 word perplexity on Wikitext with an SNN-based model on the SpiNNaker2 neuromorphic chip. Our MMF baseline model achieves 44.5 word perplexity, demonstrating potential for scalable and energy-efficient NLP applications. 

 % The original MatMul-free paper demonstrated acceleration on a D5005 Sratix 10 PAC FPGA, though the implementation was unoptimized in that digital signal processing (DSP) units were unused and burst-mode DDR memory was also unused. 
% The Loihi 2 platform overcomes several of these limitations by offering native support for sparse computation and low-power neuromorphic workloads.
% The Loihi 2 platform's optimizations for sparse processing, highly memory-efficient encoding for 1 to 8 bit weights and compute-memory integration give rise to low-power and high performance.

% \textcolor{red}{better outline neuromorphic principles to relate to the title}

\section{Model Architecture}

\begin{figure*}[ht!]
\centering
\includegraphics[width=0.9\textwidth]{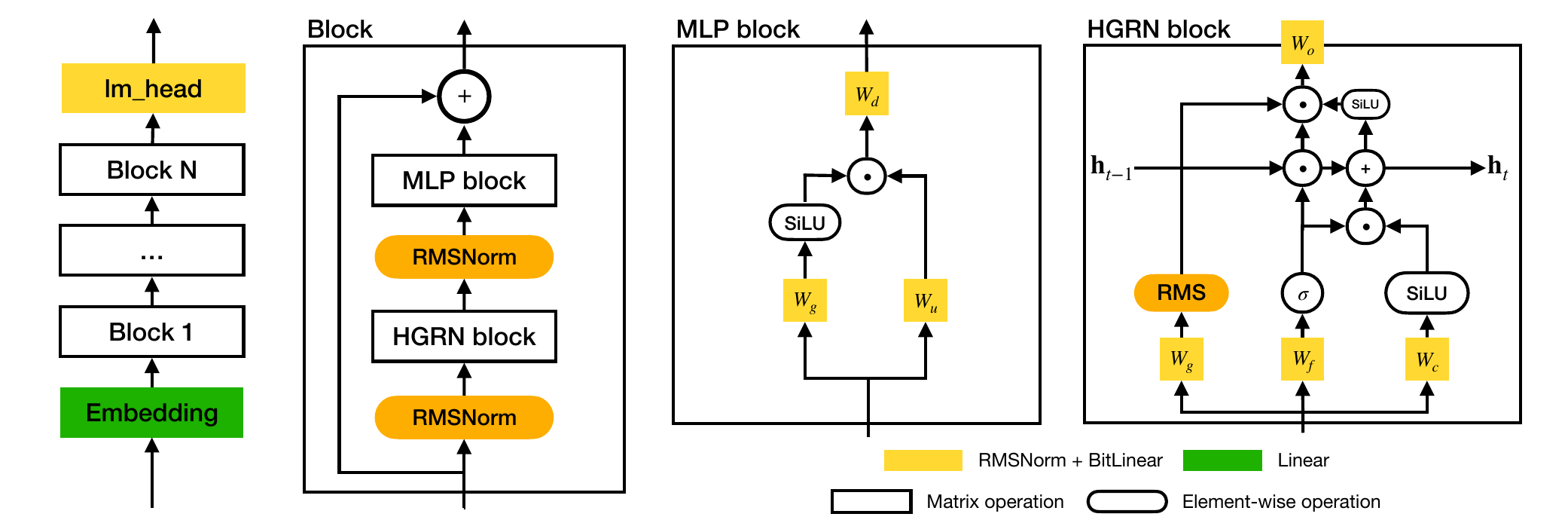}
\caption{Model architecture of the MatMul-free language model from \cite{zhu_scalable_2024}.}
\label{fig:architecture}
\end{figure*}

The model architecture is based on the 370M parameter\footnote{Available on HuggingFace: \href{https://huggingface.co/ridger/MMfreeLM-370M}{huggingface.co/ridger/MMfreeLM-370M}.} MatMul-free language model \citep{zhu_scalable_2024}. It uses a combination of ternary weights and specialized layers to replace all matrix multiplications with additions, bit shifts, and elementwise operators. 
The overall model architecture follows the Metaformer \citep{Yu_2022_CVPR} paradigm, consisting of alternating token mixers and channel mixers. Figure \ref{fig:architecture} provides a high-level overview of this structure.

The model introduces two key innovations. The \textbf{BitLinear Layer} combines a ternarized linear transformation with a preceding RMSNorm operation, to stabilize the activation distribution \citep{ma_era_2024,zhang_binarized_2023,zhu_scalable_2024}. 
Formally, the input vector $x \in \mathbb{R}^d$ (where $d=1,024$ for the 370M model), $\mathbf{W} \in \{-c, 0, c\}^{d \times d'}$ is the ternary weight matrix with a scaling factor $c\in \mathbb{R}$.
The RMSNorm \citep{zhang_root_2019} and subsequent BitLinear layer are implemented as:
\begin{align}
\text{RMSNorm}(x; g, \epsilon) &= \frac{x}{\sqrt{\epsilon + \sum_i^{d} x_i^2}} \odot g \\
\text{BitLinear}(x; \mathbf{W}, g, \epsilon) &= \text{RMSNorm}(x;g,\epsilon) \circledast \mathbf{W}
\label{eq:rmsnorm-bitlinear}
\end{align}
where $\odot$ denotes element-wise multiplication, $\epsilon=10^{-6}$ is a small constant, $g \in \mathbb{R}^{d}$ is a learned scaling parameter, and $\circledast$ performs the accumulation of ternary weights and inputs\footnote{As shown by \cite{kosson_multiplication-free_2023}, this can be done using only additions and negations.}. The use of ternary weights naturally leads to synaptic sparsity; the 370M MMF model has $35.4\% \pm 2.5\%$ weights of magnitude zero across all ternary weight matrices\footnote{Note that this synaptic sparsity does not yield memory savings because sparse encoding of a $1024\times1024$ ternary matrix takes $\approx 7.5 \times$ more memory than dense encoding. However, zero weights do allow for energy savings by skipping calculations.}.

% Notes on weight sparsity. 
% \begin{verbatim}
% bits_sparse = (1-sparsity)*N*(idx_bits+w_bits)
% bits_dense = N*w_bits
% N=1024*1024=1M
% sparsity=0.35
% idx_bits=log2(1M)=20
% bits_sparse = 15M
% bits_dense = 2M
% \end{verbatim}

The MatMul-free language model further uses a BitLinear version of the GLU (Gated Linear Unit \citep{dauphin_language_2017}):
\begin{align*}
g_t = x_t \circledast \mathbf{W}_g, \quad
u_t = x_t \circledast \mathbf{W}_u, \quad
p_t = \tau(g_t) \odot u_t, \quad
d_t = p_t \circledast \mathbf{W}_d
\end{align*}
where $g_t,u_t,p_t \in \mathbb{R}^d$ are intermediate activations, and $\tau$ denotes the SiLU activation function $\tau(x)=x \odot \sigma(x)$ where $\sigma(x) = 1/(1+e^x)$ is the sigmoid function. % and $\circledast$ represents ternary accumulation.

The \textbf{MLGRU} (MatMul-free Linear Gated Recurrent Unit) acts as the token mixer, replacing the computationally expensive self-attention mechanism in traditional transformers. It utilizes a variant of the GRU \citep{cho_learning_2014}, inspired by the HGRN \citep{qin_hierarchically_2023,qin_hgrn2_2024}, modified to use only additions and element-wise products. This is achieved by employing BitLinear layers for all linear transformations within the MLGRU cell.
Formally, we denote $h_{t} \in \mathbb{R}^d$ as the hidden state at time $t$ and $x_t \in \mathbb{R}^d$ as the input at time $t$. The dynamics of the MLGRU layer at time step $t$ are given by:
\begin{align*}
f_t &= \sigma \left( \text{BitLinear}(x_t; \mathbf{W}_f, g_f, \epsilon) \right), \\
c_t &= \tau \left( \text{BitLinear}(x_t; \mathbf{W}_c, g_c, \epsilon) \right), \\
h_t &= f_t \odot h_{t-1} + (1 - f_t) \odot c_t, \\
g_t &= \text{RMSNorm} \left( \text{BitLinear}(x_t; \mathbf{W}_g, g_g, \epsilon); g_{g'}, \epsilon \right), \\
o'_t &= g_t \odot \tau(h_t), \\
o_t &= \text{BitLinear}(o'_t; \mathbf{W}_o, g_o, \epsilon)
\end{align*}
where $\mathbf{W}_c, \mathbf{W}_f, \mathbf{W}_o, \mathbf{W}_g \in \mathbb{R}^{d \times d}$ are ternary weights, $f_t, g_t, c_t, o'_t$ are intermediate activations at time step $t$, $h_t$ is the hidden state, and $o_t$ is the final output at time step $t$. The initial hidden state $h_0$ is set to zero. Similarly to the HGRN \citep{qin_hierarchically_2023}, the MLGRU also employs the \texttt{cummax} operation to bias the forget gate values in deeper layers closer to 1, which we omit for brevity.

\section{Model Adaptation for Loihi 2}

\paragraph{Quantization of weights and activations}
As a first step, we quantize the original half-precision model from \cite{zhu_scalable_2024}. Loihi 2 supports 8-bit weights and 24-bit activations. As done by previous work \citep{ma_era_2024,zhu_scalable_2024}, we evaluate the zero-shot performance of our quantized model on a range of language tasks, including ARC (Easy \& Challenge) \citep{clark_think_2018}, Hellaswag \citep{zellers_hellaswag_2019}, Winogrande \citep{sakaguchi_winogrande_2021}, PIQA \citep{bisk_piqa_2020}, and OpenbookQA \citep{mihaylov_can_2018}.
We report the baseline models from \cite{zhu_scalable_2024} for the MatMul-free LLM that we adopt, and their transformer baseline. For comparison, we also report performance of the Qwen-2.5 500M model from \cite{qwen2.5}. Although the model has only 35\% more parameters than the transformer baseline, it performs significantly better. We expect that the baselines from \cite{zhu_scalable_2024} could reach similar performance with the training procedure from \cite{qwen2.5}.

\begin{table*}[h]
\caption{Results from quantization of the 370M MatMul-free language model on GPU. \textit{Baseline}: optimized models from \cite{zhu_scalable_2024} and \cite{qwen2.5}. \textit{PT}: PyTorch-only implementation. \textit{Ax / Wx}: activations / RMSNorm weights quantized to x-bit integers. $\epsilon_\mathrm{rms} \uparrow$: setting the value for $\epsilon_\mathrm{rms}$ to $10^{-3}$ from previously $\epsilon_\mathrm{rms}=10^{-6}$.
$^{\dagger}$: difference relative to MatMul-free baseline.
}
\centering
\begin{tabular}{l|cccccc|cr}
\toprule
Configuration                 & ARCc & ARCe & HS   & OQA  & PQ   & WG   & Avg  & Diff.$^{\dagger}$ \\
\midrule
MatMul-free baseline          & 22.8 & 42.1 & 32.4 & 28.4 & 62.6 & 49.4 & 39.6 & (0.0\%) \\
\textit{Transformer baseline} & 24.0 & 45.0 & 34.3 & 29.2 & 64.0 & 49.9 & 41.1 & (3.8\%) \\
\textit{Qwen2-500M}            & 31.0 & 64.6 & 49.1 & 35.2 & 70.3 & 56.5 & 51.1 & (29.0\%) \\
% MMF \citep{zhu_scalable_2024} & 23.8 & 42.6 & 32.8 & 28.4 & 63.0 & 49.2 & 40.3 & (1.74) \\
\midrule
PT & 22.7 & 42.2 & 32.5 & 28.4 & 62.4 & 48.5 & 39.4 & -0.4\% \\
%%% WEIGHT ONLY
PT + W8 & 23.2 & 41.8 & 32.4 & 28.0 & 62.4 & 49.5 & 39.5 &  -0.2\% \\
%%% ACTIVATION ONLY
PT + A8 & 22.7 & 40.0 & 31.5 & 27.6 & 61.0 & 50.0 & 38.8 &  -2.0\% \\
PT + A16 & 22.7 & 42.5 & 32.5 & 29.0 & 63.2 & 49.9 & 40.0 & \textbf{0.9\%} \\
\midrule
%%% ACTIVATION AND WEIGHT
PT + W8A8 & 22.3 & 40.3 & 31.9 & 27.2 & 59.9 & 49.1 & 38.5& -2.9\% \\
PT + W8A16 & 22.7 & 42.3 & 32.3 & 28.0 & 63.1 & 49.3 & 39.6 & 0.0\% \\
%%% ACTIVATION AND WEIGHT ++ epsilon
PT + W8A8 + $\epsilon_\mathrm{rms} \uparrow$ & 28.3 & 26.8 & 26.1 & 27.0 & 52.7 & 51.5 & 35.4 & -10.7\% \\
PT + W8A16 + $\epsilon_\mathrm{rms} \uparrow$ & 23.0 & 42.4 & 32.4 & 27.8 & 63.0 & 50.1 & 39.8\% & \textbf{\underline{0.4\%}} \\
\bottomrule
\end{tabular}
\label{tab:pytorch-mismatch-results}
\end{table*}

We re-implemented the model and replace the GPU-optimized Triton kernels with simple PyTorch operations that are easier to quantize and to verify against our Loihi 2 hardware implementation. Minor numerical differences add up to a relative performance drop of 0.4\%, see Table \ref{tab:pytorch-mismatch-results} (MatMul-free baseline vs. PyTorch).
We further quantize weights and activations, to ensure compatibility with fixed point computation, as required by Loihi 2. Quantization is applied symmetrically per tensor, with scaling factors restricted to powers of two so that rescaling can be done efficiently using bit-shift operations. 
% Formally, let $x \in \mathbb{R}^d$ be a vector of activations or element-wise normalization weights (all weight \textit{matrices} are ternary), then quantization is performed as:
% \begin{align}
% \hat{x} = \text{round}\left(\frac{x}{s_x}\right), \quad s_x = 2^{\lfloor \log_2(\max(|x|)) \rceil}
% \end{align}
% where $\lfloor \cdot \rceil$ denotes the rounding operation, and $\max(|x|)$ is the absolute maximum activation observed on the data across all activation channels. 
We use ``fake quantization'' in PyTorch, in that all operations use floating-point numbers which are quantized and de-quantized.

Our results in Table \ref{tab:pytorch-mismatch-results} show that 8-bit weight quantization leads to only a $0.2\%$ performance decrease relative to the baseline model. 
Previous work has demonstrated low quantization errors for state space models when using W8A16 (8-bit weights, 16-bit activations) with significantly higher drops for W8A8 \citep{pierro_mamba-ptq_2024,abreu_q-s5_2024,chiang_quamba_2024}. Indeed, quantization to W8A8 and W8A16 of our model show a relative performance drop of 2.9\% and 0.0\%, respectively. We thus use W8A16 for our hardware implementation.

Although the baseline model uses half precision (FP16), the RMSNorm is still computed in full precision (FP32) for numerical stability. We quantize activations inside the RMSNorm layer to 24-bit integers with 12 fractional bits. We further increase the $\epsilon$ value from $10^{-6}$ (which underflows with 12 fractional bits) to $10^{-3}$. The resulting performance of all interventions mentioned thus far is shown in the last two rows of Table \ref{tab:pytorch-mismatch-results}. The W8A8 and W8A16 quantization schemes with modified $\epsilon$ show a relative performance change of -10.7\% and +0.4\%, respectively. We chose the W8A16 quantization scheme as it is fully compatible with Loihi 2. Therefore, our final quantized model on GPU shows no performance loss compared to the baseline FP16 model.

\paragraph{Fixed-point implementation}\label{ss:fxp-implementation}
Two operations used in the MatMul-free LM are not defined on integers, namely the sigmoid activation function $\sigma$ and the inverse-square-root in the RMSNorm. 
We employ a look-up-table (LUT) for a fixed-point approximation of the logistic sigmoid function, $\sigma(x) = 1/(1 + e^{-x})$. For the inverse-square-root in the RMSNorm layer, we adapted a well-known ``fast inverse square root'' algorithm to operate in fixed-point arithmetic. See Appendix \ref{app:fxp-implementation} for details.

\paragraph{Mapping the model to Loihi 2}
\label{ss:mapping-model-to-l2}

The Loihi 2 implementation represents the model as a network of neurons interconnected via synapses. Each neuron is implemented as a simple microcode program that is executed asynchronously on one of 120 neuro cores on each Loihi 2 chip, after which its output is transmitted to other neurons through synaptic connections. A global time step is maintained through barrier synchronization between all neuro cores.
Since each neuron only maintains its own state, aggregate operations--such as computing the sum of squares over an activation vector--must be realized by dedicated neurons that receive inputs from all neurons within the corresponding layer. This is the case for the RMSNorm, where the sum of squares over all neurons in a given layer is calculated, see Equation \ref{eq:rmsnorm-bitlinear}. Figure \ref{app:nxkernel-graph} (\textit{right}) illustrates the computational graph that implements the RMSNorm operation on Loihi 2. 
Layer and operator fusion are well-established strategies to minimize redundant computations \citep{waeijen_convfusion_2021,niu_dnnfusion_2021}, thereby reducing latency and enhancing energy efficiency. In our implementation, we perform Loihi-specific layer fusion to consolidate operations into a streamlined computational graph, as depicted in Figure \ref{app:nxkernel-graph} (\textit{left}). We also derive the fusion of two subsequent RMSNorm layers into a single operator for further acceleration, see Appendix \ref{app:double-rmsnorm}.

\section{Results}
\label{sec:results}

We implemented a single block of the MatMul-free language model on a single Loihi 2 chip. We parallelize the workload to all 120 available neuro cores on the chip. Verification of the model on Loihi 2 indicates close alignment with the quantized PyTorch simulation. 

\noindent\textbf{Note: All current comparisons are performed with FP16 baselines on non-Loihi hardware.}

We contrast the estimated performance of the 370M MatMul-free model on Loihi 2 against transformer-based baselines running on an NVIDIA Jetson Orin Nano. We selected the Jetson Orin Nano (8GB) as our comparison platform because it represents a state-of-the-art edge AI device with 1024 CUDA cores, 32 Tensor cores, and a maximum power consumption of 15W, making it a relevant benchmark for energy-efficient AI applications. The Orin Nano is designed specifically for edge deployment scenarios similar to those targeted by neuromorphic hardware, enabling a fair comparison between platforms intended for similar operational environments. 
Efficiency and throughput metrics for Loihi 2 are estimates based on a preliminary implementation that is not fully optimized, see Appendix \ref{app:l2-results-detailed} for details.
We compare the MatMul-free LLM on Loihi 2 against two similarly-sized transformer-based LLMs available on HuggingFace.
As a Llama series representative model, we use Alireo-400M \citep{alireo2024}, a 400M parameter transformer-based LLM with 24 layers and a context window of 8,192. It should be noted that the Alireo model was trained specifically on Italian text, so its performance is not included in Table~\ref{tab:pytorch-mismatch-results} as it would not be representative of a competitive general-purpose transformer-based model at this parameter scale. 
We further use Qwen2.5-500M \citep{qwen2,qwen2.5}, a 500M parameter transformer-based LLM with 24 layers and a context window of 32,768 whose performance is also included in Table~\ref{tab:pytorch-mismatch-results}.
Both models run in half-precision (FP16).
We did not benchmark the MatMul-free LLM or the Transformer baseline from \cite{zhu_scalable_2024} because the Jetson Orin Nano does not support Triton.

Table \ref{tab:jetson-loihi-comp} shows results for the comparison of the MatMul-free LLM on Loihi 2 and transformer-based LLMs on the NVIDIA Jetson Orin Nano, also including results for the MatMul-free LLM on an H100 GPU and the Transformer++ baseline from \cite{zhu_scalable_2024} on a single H100 GPU. The results demonstrate that the MatMul-free LLM on GPU improves in throughput and efficiency with longer sequence lengths, due to linear scaling of the token mixing and better utilization of the GPU at higher sequence lengths. In contrast, the transformer++ baseline on GPU increases in throughput for short sequence lengths due to better utilization of the hardware, and then deteriorates in throughput and efficiency for even longer sequences because of the quadratic scaling of self-attention. 

\begin{table*}[h]
\caption{
Throughput and energy efficiency for two transformer-based language models running on the NVIDIA Jetson Orin Nano and H100 compared to our MatMul-free LM running on Intel's Loihi 2, across different sequence lengths for prefill and generation. 
The best-performing sequence length for each model and metric is \underline{underlined}.
Metrics for Loihi 2 are based on preliminary experiments and subject to further performance optimization, see Appendix \ref{app:l2-results-detailed}. 
\textbf{Gen}: autoregressive generation, \textbf{Prefill}: prefill mode. 
$^*$ Llama representative model from \cite{alireo2024}.
}
\centering

%%%%%%%%%%%%%%%%%%%%%%%%%%%%%%%%%%%%%%%%%%%%%%%%%%%%%%%%%%%%%%%%%%%%%%%%%%%%%%%%%%%%
%%%%%%%%%%%%%%%%%%%%%%%%%%%%%%%%%%%%%%%%%%%%%%%%%%%%%%%%%%%%%%%%%%%%%%%%%%%%%%%%%%%%

\begin{tabular}{clr|rrrrr|rrrrr}
\toprule
 & & & \multicolumn{5}{c}{Throughput ($\uparrow$ tokens/sec)} & \multicolumn{5}{c}{Efficiency ($\downarrow$ mJ/token)} \\
 & \multicolumn{2}{r}{Sequence length} & 500 & 1000 & 4000 & 8000 & 16000 & 500 & 1000 & 4000 & 8000 & 16000 \\

\midrule
\multirow{5}{*}{\rotatebox{90}{Generate}}
 & \textbf{MMF (370M)} & \textbf{Loihi 2$^\dagger$} & \textbf{41.5} & \textbf{41.5} & \textbf{41.5} & \textbf{41.5} & \textbf{41.5} & \textbf{405} & \textbf{405} & \textbf{405} & \textbf{405} & \textbf{405} \\
 & MMF (370M)          & H100     & 13.4       & 13.3 & \underline{13.5}    &  13.2   & \underline{13.5}  &  10.1k   & 10.1k & 10.0k    & 9.9k   &  \underline{9.8k} \\
 & TF++ (370M)          & H100    & 22.4       & \underline{22.9} & 21.7    &  21.3   & 20.9  &  \underline{5.5k}   & 5.6k  & 6.2k    & 6.8k   &  8.2k \\
 & Llama$^*$ (400M)     & Jetson$^\ddagger$  & 14.3    & 14.9 & 14.7 & \underline{15.2} & 12.8  & 723 & \underline{719}   & 853  & 812 & 1.2k \\
 & Qwen2 (500M)         & Jetson$^\ddagger$  & 13.4    & 14.0 & 14.1 & \underline{15.4} &  12.6     & 791 & \underline{785}   & 912  & 839 &  1.2k   \\

\midrule
\multirow{5}{*}{\rotatebox{90}{Prefill}}
 & \textbf{MMF (370M)} & \textbf{Loihi 2$^\dagger$} & \textbf{6632} & \textbf{6632} & \textbf{6632} & \textbf{6632} & \textbf{6632} & \textbf{3.7} & \textbf{3.7} & \textbf{3.7} & \textbf{3.7} & \textbf{3.7} \\
 & MMF (370M)          & H100     & 11.4k    & 13.1k & 30.6k    &  51.6k   & 84.6k  & 6.1     & 5.3  & 2.5    & 1.4    & 0.9  \\
 & TF++ (370M)          & H100    & 21.6k    & 32.7k & 44.3k     & 55.4k   & 60.5k    & 11.3     & 7.3  & 5.4    & 4.3    & 3.8  \\
 & Llama$^*$ (400M)     & Jetson$^\ddagger$  & 849  & 1620  & \underline{3153} & 2258 & 1440   & 11.7  & 7.8  & \underline{6.8}  & 7.6  & 11.5  \\
 & Qwen2 (500M)         & Jetson$^\ddagger$  & 627  & 909   & 2639 & \underline{3861} & 3617   & 17.9  & 13.9 & 6.7  & \underline{4.4}  & 5.3 \\
\bottomrule
\multicolumn{13}{p{12.5cm}}{
\tiny$^\dagger$ The MatMul-free LM on Loihi 2 was characterized on a 32-chip Alia Point Loihi 2 system (N3C1 silicon) running NxKernel v0.2.0 and NxCore v2.5.8 (accessible to Intel Neuromorphic Research Community members). Appendix \ref{app:l2-results-detailed} compares results for single-chip and multi-chip scaling.
\par
$^\ddagger$ Transformer LMs were characterized on NVIDIA Jetson Orin Nano 8GB using the MAXN power mode running Jetpack 6.2, TensorRT 10.3.0, CUDA 12.4. Energy values include CPU\_GPU\_CV, SOC, and IO components as reported by jtop 4.3.0.
\par
Performance results are based on testing as of Jan 2025 and may not reflect all publicly available security updates. Results may vary.
}
\end{tabular}
\label{tab:jetson-loihi-comp}
\end{table*}

In our experiments, we focus on single-batch inference and we further differentiate between two operational modes. ``Prefill'' refers to the phase where a long input sequence is ingested, allowing for parallel processing of multiple tokens, which naturally yields higher throughput and energy efficiency. In contrast, ``autoregressive generation'' denotes the sequential production of tokens, where each token must be generated and received before the next can be processed. Profiling both modes separately highlights these differences in performance and efficiency. The execution modes that enable this distinct processing on Loihi 2 are described in Appendix \ref{app:exmode}.

During \textit{prefill}, Loihi 2 shows at least \textbf{$\mathbf{2\times}$  higher throughput} with approximately \textbf{$\mathbf{2\times}$  less energy} per token.
During \textit{auto-regressive generation}, the advantage of Loihi 2 grows to having almost \textbf{$\mathbf{3\times}$  higher throughput} with approximately \textbf{$\mathbf{2\times}$ less energy} per token. Due to the MatMul-free model's subquadratic architecture, this advantage is expected to grow for longer sequence lengths.
We highlight that power and throughput of the MatMul-free model on Loihi 2 is \textit{constant across sequence length}. 
This is due to the linear scaling of the recurrent token mixer and the fact that Loihi 2 has its parameters and hidden states stored locally inside neuro-cores thus requiring significantly less memory movement than GPUs. 
Further optimizations are underway to increase throughput and reduce power consumption of the MatMul-free language model on Loihi 2, see Appendix \ref{app:detailed-hw-results}.

Table \ref{tab:jetson-loihi-comp} highlights how transformers on an edge GPU are consistently slower and less efficient during generation, and scale unfavorably to longer sequence lengths. 
During prefill, transformers show low throughput and efficiency for shorter sequences due to under-utilization of the GPU, then reach optimal performance between 4000-8000 tokens, after which both throughput and efficiency decline. 
Table \ref{tab:jetson-loihi-comp} also compares the MatMul-free model on Loihi against the same model running on an H100 GPU. Loihi 2 delivers \textbf{$\mathbf{3\times}$ higher throughput} during generation with at least \textbf{$\mathbf{14\times}$ less energy per token}. During prefill, the H100 delivers higher throughput than Loihi 2, but outperforms in energy efficiency only for large sequence lengths.
% \textcolor{red}{also discuss latency/time-to-first-token}

Latency characteristics are particularly important for interactive applications at the edge. Our experiments reveal that for batch size 1, which is typical in edge deployment scenarios, the MatMul-free model on Loihi 2 demonstrates significant latency advantages, with a \textbf{$6.6 \times$ lower time-to-first token} on a 500-token input sequence (99ms for the MatMul-free model on the Loihi 2 vs. 659ms for the Llama-style model on the Jetson). 
% TTFT 500-input: 99ms on L2  --     1/6632*500+1/41,5
% TTFT 500-input: 659ms on Jetson -- 1/6632*500+1/41,5
This advantage increases with sequence length due to the linear scaling properties of our approach versus the quadratic complexity of transformer models. For real-time applications like voice assistants or mobile chatbots, this latency reduction directly translates to more responsive user experiences while maintaining significantly lower energy consumption. 

While we benchmarked against the Jetson Orin Nano as a representative edge GPU platform, we expect our approach to show similar advantages compared to other edge computing solutions. Platforms such as mobile SoCs, FPGAs, and various edge TPUs all face similar challenges with transformer-based models: quadratic scaling with sequence length and significant memory movement costs. Since our neuromorphic approach addresses both fundamental limitations through linear scaling and closer compute-memory integration, we anticipate the relative throughput and energy efficiency advantages to persist across other edge computing architectures. The benefits would be most pronounced for resource-constrained IoT or mobile platforms where energy efficiency is paramount, with potentially more modest gains against specialized NPUs that have already optimized for recurrent operations.

% Initial experiments for scaling the MatMul-free language model to multiple Loihi 2 chips showed that the inter-chip communication is low because only one activation vector must be transmitted from one chip to the next at any time step. We characterized an initial slowdown by $k_\text{comm} \approx \ 21\%$ in the processing. This would reduce our throughput in prefill mode by $k_\text{comm}$ and in generation mode by $k_\text{comm}/27 \approx 1\%$ because only every 27 time steps there will be an activation communicated across chips. We thus expect the MatMul-free LLM to scale favorably to multiple chips in order to run inference on the full model at the throughput and energy that is presented in Table \ref{tab:jetson-loihi-comp}.

\section{Conclusion}

We demonstrated that neuromorphic principles and co-design can be leveraged to build efficient LLMs. By merging a MatMul‐free architecture with Intel's Loihi 2--leveraging state-based computation, extreme quantization, and operator fusion--we built a strong 370M-parameter model with significantly improved throughput and energy efficiency. Our experiments reveal that the inherent parallelism and low-power processing of Loihi 2 translate into substantial gains in throughput and energy efficiency.

The key innovations of our approach include: (1) the first demonstration of a modern LLM architecture on neuromorphic hardware, establishing a pathway for efficient AI at the edge; (2) a hardware-aware quantization methodology that maintains model accuracy while enabling fixed-point computation; (3) a novel microcode implementation of the MatMul-free architecture that exploits Loihi 2's asynchronous, event-driven computing paradigm; and (4) custom operator fusion techniques specifically designed for neuromorphic computation, including our double RMSNorm derivation. Unlike previous approaches that targeted specific neural primitives for neuromorphic systems, our work shows that complete, competitive language models can be adapted to leverage the unique characteristics of neuromorphic hardware while maintaining performance.

On the hardware side, our results demonstrate the potential of neuromorphic processors as platforms for scalable efficient inference, suggesting that future architectures can be co-designed with model innovations to further push performance limits. 
Our approach offers a promising pathway to enable adaptive language processing without the prohibitive energy costs associated with traditional LLMs. 
Given the rising importance of reasoning models that require extended chain-of-thought rollouts \citep{deepseek-ai_deepseek-r1_2025}, efficient and high-throughput autoregressive generation is more critical than ever. Our design excels especially in this mode, paving the way for scalable foundation models capable of reasoning faster and more efficiently.

\bibliography{iclr2025_conference}
\bibliographystyle{iclr2025_conference}

\newpage
\appendix
\section{Appendix}

\subsection{Loihi 2 Hardware Architecture}
\label{app:l2-hw-arch}

\begin{figure}[h!]
    \centering
    \includegraphics[width=0.45\linewidth]{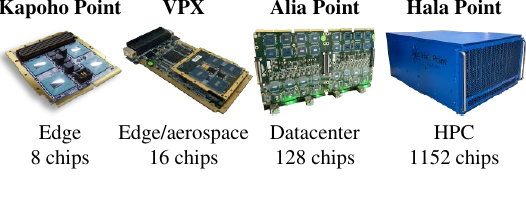}
    \caption{Different Loihi 2 systems are available to cover a wide range of applications from the edge to HPC with up to 1 billion neurons.}
    \label{fig:loihi_systems}
\end{figure}

Loihi 2 is the second-generation of Intel's neuromorphic research processor that was designed for sparse, event-based neural networks \citep{orchard2021efficient}.
On the Loihi 2 chip, a neural network is processed by massively parallel compute units called \textit{neuro-cores}, with 120 such neuro-cores per chip. Multiple Loihi 2 chips can be stacked together into various larger systems with up to 1,152 chips, see Figure \ref{fig:loihi_systems}.
The neuro-cores compute and communicate asynchronously, but a global algorithmic time step is maintained through barrier synchronization. The neuro-cores are co-located with memory and can thus efficiently update local states, simulating up to 8192 stateful neurons per core. Each neuron can be programmed by the user to realize a variety of temporal dynamics through assembly code, and can use a variable amount of memory--each neuro-core has a fixed amount of memory but one can implement neurons with more memory by trading off the number of neurons that each core implements.
Input from and output to external hosts and sensors is provided with up to 160 million 32-bit integer messages per second \citep{shrestha_efficient_2024}.
Loihi 2 can scale to real-world workloads of various sizes with up to 1 billion neurons and 128 billion synapses, using fully-digital stacked systems (Hala Point, Figure \ref{fig:loihi_systems}).

The architectural features of Loihi 2 offer unique opportunities to compress and optimize deep learning models. Neural networks running on Loihi must be quantized to low-precision using fixed-point arithmetic--8 bits for synaptic weights\footnote{This is not a hard limit, as one can implement an $8n$-bit synapse through $n$ separate 8-bit synapses that are added together with different fixed-point exponents.} and up to 32 bits for messages\footnote{Local states are not restricted in precision, and one may also transmit messages with more than 32 bits in an analogous way to what is described above for synaptic weights.}. Unlike GPUs, Loihi 2 is optimized for computations local within neurons, a common focus of neuromorphic processors.
First, it allows fast and efficient updates of neuronal states with recurrent dynamics with minimal data movement, due to the local memory of each neuro-core.
Second, the asynchronous event-driven architecture of Loihi 2 allows it to efficiently process unstructured sparse weight matrices.
Third, the neuro cores can leverage sparsified activation between neurons, as the asynchronous communication transfers only non-zero messages.

\subsubsection{Execution Modes on Loihi 2}
\label{app:exmode}

\begin{figure}[h]
    \centering
    \includegraphics[width=0.5\linewidth]{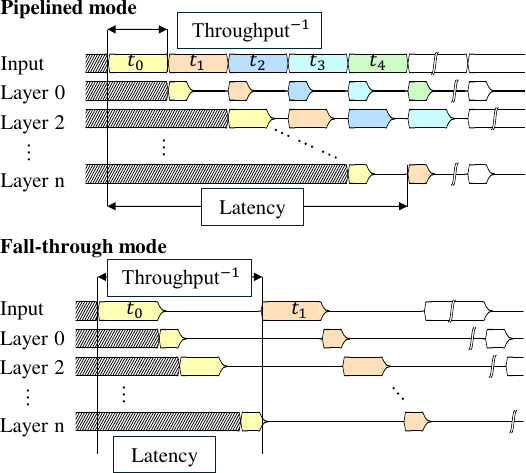}
    \caption{Different execution modes on Loihi 2 that either optimize throughput or latency. In the \textit{pipelined mode}, a new data point is inserted in each time step, to use all processing cores and maximize the throughput--at the expense of latency because equal time bins $t_0=t_1=\ldots$ are enforced. In the \textit{fall-through mode}, a new data points is only provided once the last data point has been fully processed with minimum latency. Only a single neuronal layer is active at any step as data travels through the network. The time per step is thus minimized as traffic is reduced and potentially more complex neuronal layers are not updated.}
    \label{fig:loihi_exec_modes}
\end{figure}

Loihi 2's asynchronous architecture enables a trade off between throughput and latency, as illustrated in Figure \ref{fig:loihi_exec_modes}. For optimal throughput, new input is provided every time step and forwarded through the neuronal layers in a \textit{pipelined mode}. For optimal latency we use \textit{fall-through mode}, where new input is injected only once the previous input has been processed by, or fallen through, the network as fast as possible. 
%The pipelined and fall-through mode can be balanced by changing the rate of new input, to match the throughput of a given input stream while minimizing its processing latency.
Naturally, when processing a long sequence of prefill text in an LLM, we use pipeline mode for optimal throughput. When generating new text in auto-regressive generation of an LLM, we have to wait for the token at time $t$ to be output before we can begin processing the next token at time $t+1$, thus making this a natural fit for Loihi 2's fall-through mode.

\subsection{Fixed-Point Implementation Details}
\label{app:fxp-implementation}

\subsubsection{Fixed-Point Implementation of the Sigmoid Function}
\label{app:fxp-sigmoid}

We employ a look-up-table (LUT) for a fixed-point approximation of the logistic sigmoid function, $\sigma(x) = 1/(1 + e^{-x})$, as discussed in Section \ref{ss:fxp-implementation}. Specifically, we scale the floating-point input $x$ by $2^{x_{\text{exp}}}$ where $x_{\text{exp}}=6$ determines the accepted input precision of the fixed-point sigmoid implementation. We quantize $x 2^{x_\text{exp}}$ to an integer domain $x_\text{fxp} = \lfloor x 2^{x_\text{exp}} \rceil$, and store precomputed 
% $y_\text{int}^\text{lut,i} = \lfloor \sigma\!\bigl(\frac{x^\text{lut,i}_\text{int}}{2^{x_{\text{exp}}}}\bigr)\cdot 2^{y_{\text{exp}}}\rfloor$ 
values in a LUT $\left(x_\text{int}^\text{lut,i} \mapsto y_\text{int}^\text{lut,i}\right)_{i \in \{0, \ldots N_\sigma\}}$, where $N_\sigma=8$ determines the number of LUT entries:
\begin{equation}
    y_\text{int}^\text{lut,i} = \lfloor \sigma\!\bigl(\frac{x^\text{lut,i}_\text{int}}{2^{x_{\text{exp}}}}\bigr)\cdot 2^{y_{\text{exp}}}\rfloor
\end{equation}
This LUT stores only entries for positive inputs. For negative inputs, we exploit $\sigma(-x) = 1 - \sigma(x)$, thus requiring only half-range values. During inference, a piecewise linear interpolation between adjacent LUT entries refines the output. This design offers efficient computation and controllable approximation error by tuning $x_\text{exp}$ and $N_\sigma$.

\subsubsection{Fixed point implementation of the inverse square root}
\label{app:fixed-pt-invsqrt}

For the inverse-square-root in the RMSNorm layer, we adapted a well-known ``fast inverse square root'' algorithm \texttt{FastInvSqrt} to operate entirely in fixed-point arithmetic, as discussed in Section \ref{ss:fxp-implementation}.
Our method treats the input $\tilde{x}$ as an integer paired with a fixed exponent, and uses a LUT with 24 values to produce an initial guess for $\sqrt{\tilde{x}}$. This estimate is then refined using five iterations of the Newton-Raphson method, all in a fixed-point format.

\subsection{Double RMSNorm Derivation}
\label{app:double-rmsnorm}

Let $x \in \mathbb{R}^d$ and $y=\text{RMSNorm}(x)$ and $z=\text{RMSNorm}(y)$, in expanded form:
\begin{align}
    y &= \frac{x}{\sqrt{\epsilon + \sum_i^{d} x_i^2}} \odot g_1 \\
    z &= \frac{y}{\sqrt{\epsilon + \sum_i^{d} y_i^2}} \odot g_2
\end{align}
We wish to derive an equation for $z = \text{DoubleRMSNorm} (x) = \text{RMSNorm}(\text{RMSNorm}(x))$.

First, we express $\mu_{\mathbf{y}}$ in terms of $\mu_{\mathbf{x}}$:
\begin{align}
\mu_{\mathbf{y}} &= \frac{1}{D} \sum_{i=1}^D y_i^2 = \frac{1}{D} \sum_{i=1}^D \left( x_i \cdot \frac{g_1}{\sqrt{\mu_{\mathbf{x}} + \varepsilon}} \right)^2 \\
&= \left( \frac{g_1^2}{\mu_{\mathbf{x}} + \varepsilon} \right) \cdot \left( \frac{1}{D} \sum_{i=1}^D x_i^2 \right) = \frac{g_1^2 \mu_{\mathbf{x}}}{\mu_{\mathbf{x}} + \varepsilon}.
\end{align}
We then express $\mathbf{z}$ in terms of $\mathbf{x}$ by plugging in the equation for $\mathbf{y}$:
\begin{align}
\mathbf{z} &= \mathbf{y} \cdot \frac{g_2}{\sqrt{\mu_{\mathbf{y}} + \varepsilon}} = \left( \mathbf{x} \cdot \frac{g_1}{\sqrt{\mu_{\mathbf{x}} + \varepsilon}} \right) \cdot \frac{g_2}{\sqrt{\mu_{\mathbf{y}} + \varepsilon}} \\
&= \mathbf{x} \cdot \frac{g_1 g_2}{\sqrt{ (\mu_{\mathbf{x}} + \varepsilon)(\mu_{\mathbf{y}} + \varepsilon) }}.
\end{align}
We simplify the denominator:
\begin{align}
\sqrt{ (\mu_{\mathbf{x}} + \varepsilon)(\mu_{\mathbf{y}} + \varepsilon) } &= \sqrt{ (\mu_{\mathbf{x}} + \varepsilon) \cdot \frac{g_1^2 \mu_{\mathbf{x}}}{\mu_{\mathbf{x}} + \varepsilon} + \varepsilon } \\
&= \sqrt{ (\mu_{\mathbf{x}} + \varepsilon) \cdot \frac{ \mu_{\mathbf{x}} ( g_1^2 + \varepsilon ) + \varepsilon^2 }{ \mu_{\mathbf{x}} + \varepsilon } } \\
&= \sqrt{ \mu_{\mathbf{x}} ( g_1^2 + \varepsilon ) + \varepsilon^2 }.
\end{align}
We then derive the final expression for $\mathbf{z}$:
\begin{align}
\mathbf{z} &= \mathbf{x} \cdot \frac{ g_1 g_2 }{ \sqrt{ \mu_{\mathbf{x}} ( g_1^2 + \varepsilon ) + \varepsilon^2 } }.
\end{align}

This provides the combined RMSNorm operation with different scaling parameters $g_1$ and $g_2$.
By combining two RMSNorm operations with different scaling parameters, we arrive at a single normalization step:
\begin{equation}
\mathbf{z} = \mathbf{x} \cdot \frac{ g_{\text{combined}} }{ \sqrt{ \mu_{\mathbf{x}} + \varepsilon_{\text{combined}} } },
\end{equation}
where:
\begin{align}
g_{\text{combined}} = \frac{ g_1 g_2 }{ \sqrt{ g_1^2 + \varepsilon } }, \quad 
\varepsilon_{\text{combined}} = \frac{ \varepsilon^2 }{ g_1^2 + \varepsilon }.
\end{align}

Alternatively, since the denominator depends on $\mu_{\mathbf{x}}$, it may not be possible to express $\varepsilon_{\text{combined}}$ independently without further approximations.

\subsection{Detailed Hardware Results}
\label{app:detailed-hw-results}

\subsubsection{Detailed Loihi 2 Results}
\label{app:l2-results-detailed}

\paragraph{Single chip experiments}
As described in Section \ref{sec:results}, the energy and throughput metrics for Loihi 2 were estimated based on a preliminary implementation. We first implemented a single MatMul-free LM block on a Oheo Gulch single-chip Loihi 2 system, see Table \ref{tab:loihi-table-1-24-chip} (\textit{1-chip}). We measured the average time per step (TPS, $T_\text{TPS}$), \textit{i.e.}, the time that a single execution time step takes. Given the number of time steps per block, $N_\text{steps/block}$, we can compute the total latency of the model, or the \textit{time-to-first-token} $T_\text{ttft}$, as $T_\text{ttft} = N_\text{blocks} \times N_\text{steps/block} \times T_\text{TPS}$ where $N_\text{blocks}=24$ for the 370M MatMul-free language model. 

\begin{figure}[h]
    \centering
    \includegraphics[width=0.9\linewidth]{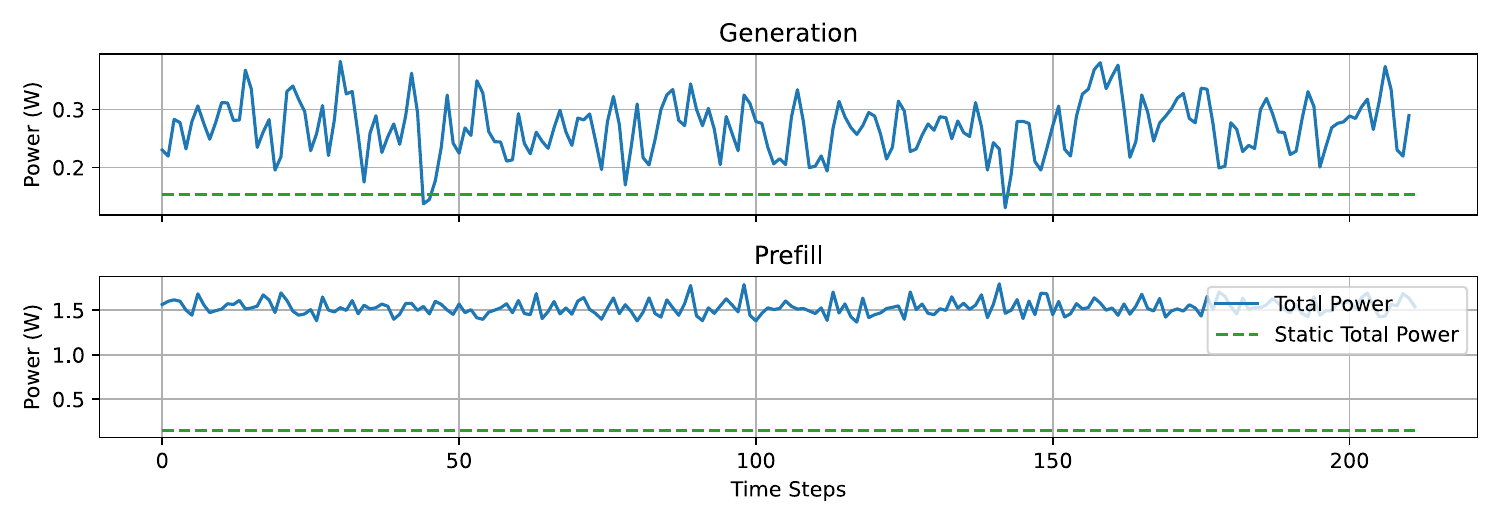}
    \caption{Power of one MatMul-free block on a single-chip Loihi 2 system.}
    \label{fig:l2-power-single-chip}
\end{figure}

In prefill we use pipelined mode where the TPS is constant over time because equal time bins are enforced (see Appendix \ref{app:exmode} for an explanation). We calculate the prefill throughput as $f_\text{prefill} = T_\text{TPS}^{-1}$ because a new token is processed in the interval $T_\text{TPS}$. We also measure the power of the single-chip system as the sum of a static power and dynamic power component: $P^\text{1-chip}=\tilde P^\text{1-chip} + \bar P^\text{1-chip}$ where $\bar P$ denotes static power and $\tilde P$ denotes dynamic power. We estimate the total prefill power as $\hat P_\text{prefill} = 24 \times P^\text{1-chip}$. We finally estimate the energy per token as $\hat E_\text{prefill} = \hat P_\text{prefill} * T_\text{TPS}$. Figure \ref{fig:l2-power-single-chip} shows the dynamic and static power of the single-chip experiment.

In generation we use fallthrough mode where the TPS varies over time and directly reflects the latency of the operation that is done at the current time step (see Appendix \ref{app:exmode} for an explanation). We average over all TPS values across at least 1000 time steps to get $T_\text{TPS}$. We then calculate the generation throughput as $f_\text{generate} = T_\text{ttft}^{-1}$. This is typically significantly lower than $f_\text{prefill}$. We measure power in the same way as in prefill mode. However, to estimate the total power for the full model with all 24 blocks, we use $\hat P_\text{generate} = \tilde P^\text{1-chip} + 24 \times \bar P^\text{1-chip}$ because at any given time only a single chip will be processing information and drawing dynamic power--all other chips will be idling and drawing only static power. We estimate the energy per token as $\hat E_\text{generate} = \hat P_\text{generate} * T_\text{ttft}$. 

\begin{table*}[h]
\caption{Throughput and energy efficiency estimates for our MatMul-free LM running on Intel's Loihi 2, based on a 1-chip implementation and a 24-chip implementation. Each chip implements a single block of the language model. \textbf{GEN}: autoregressive generation, \textbf{PF}: prefill mode.}
\centering
\begin{tabular}{cl|cc}
\toprule
 && Throughput ($\uparrow$ tokens/sec) & Efficiency ($\downarrow$ mJ/token) \\
\midrule
\multirow{2}{*}{\rotatebox{90}{GEN}} &
% ./profiling/profiling_iclr-test-store-hicore-ap-24chip-tmp-ap01.html
\textbf{Ours (370M)} (24-chip) & \textbf{41.5} & \textbf{405} \\
& \textbf{Ours (370M)} (1-chip) & \textbf{71.3} & \textbf{59} \\
\midrule
\multirow{2}{*}{\rotatebox{90}{PF}} &
\textbf{Ours (370M)} (24-chip) & \textbf{6632} & \textbf{3.7} \\
& \textbf{Ours (370M)} (1-chip) & \textbf{13965} & \textbf{2.8} \\
\bottomrule
\multicolumn{4}{p{12.5cm}}{\tiny$^*$ The MatMul-free LM on Loihi 2 was characterized on (1) Oheo Gulch single-chip Loihi 2 system, (2) Alia Point 32-chip Loihi 2 system (both: N3C1 silicon) running NxKernel v0.2.0 and NxCore v2.5.8 (accessible to Intel Neuromorphic Research Community members). Detailed results for single-chip and multi-chip scaling are presented in Appendix \ref{app:l2-results-detailed}.
\par
$^\ddagger$ Transformer LMs were characterized on NVIDIA Jetson Orin Nano 8GB using the MAXN power mode running Jetpack 6.2, TensorRT 10.3.0, CUDA 12.4. Energy values include CPU\_GPU\_CV, SOC, and IO components as reported by jtop 4.3.0.
\par
Performance results are based on testing as of Jan 2025 and may not reflect all publicly available security updates. Results may vary.
}
\end{tabular}
\label{tab:loihi-table-1-24-chip}
\end{table*}

\paragraph{Multi-chip experiments}
Naturally, our estimates based on single-chip experiments ignore additional latency and power that comes from inter-chip communication. Therefore, we implement all 24 blocks of the MatMul-free LM on a larger multi-chip system. We use the Alia Point system (see Figure \ref{fig:loihi_systems}), where we run only 32 of all 128 chips. Each block of the MatMul-free LM is mapped to exactly one chip, thus we run the entire model on 24 out of the 32 chips on our Alia Point system. We calculate the throughput as for the single-chip experiments, but power and energy per token are now measured directly, rather than being estimates. Table \ref{tab:loihi-table-1-24-chip} shows the comparison of our single-chip estimates against the full 24-chip implementation of all MatMul-free blocks on a 32-chip Alia Point system. 

\begin{figure}[h]
    \centering
    \includegraphics[width=0.8\linewidth]{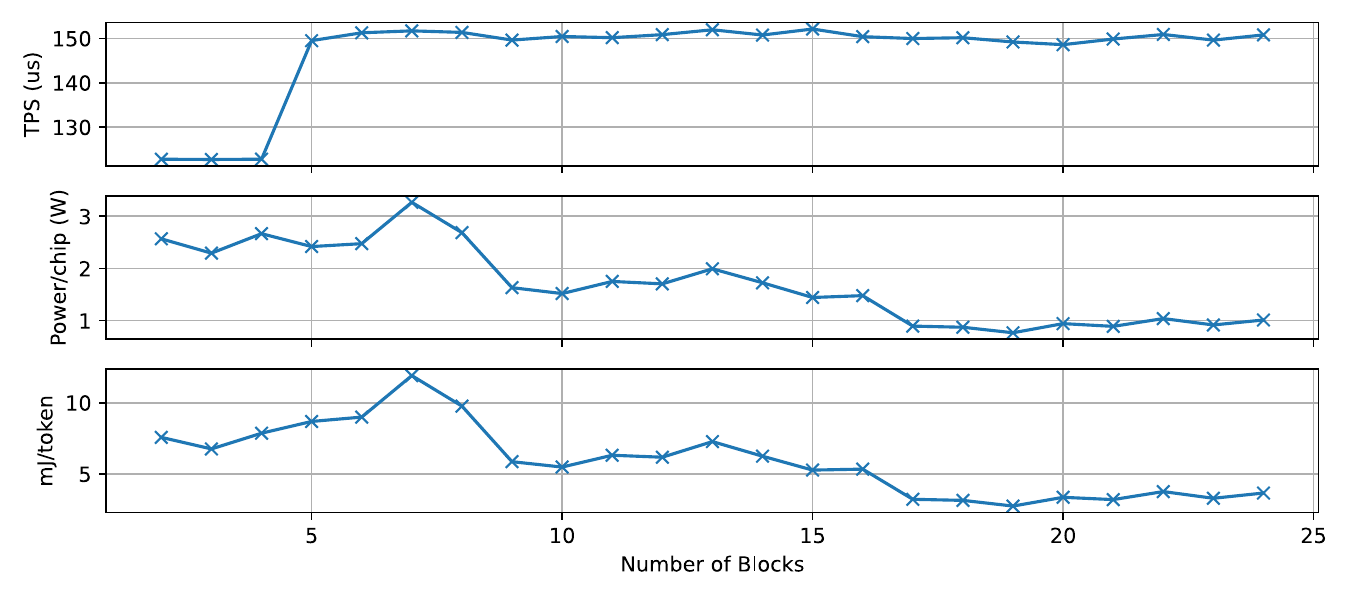}
    \caption{Scaling of time per step (inversely proportional to throughput, see text), power per chip and energy per token, as more chips are utilized in a 32-chip Alia Point Loihi 2 system. Each block of the MatMul-free LLM is implemented on a single Loihi 2 chip.}
    \label{fig:l2-chip-scaling}
\end{figure}

Throughput is reduced by $\approx 2.1 \times$ for prefill and by $\approx 1.7 \times$ for generation. Part of this slowdown comes from inter-chip communication. Figure \ref{fig:l2-chip-scaling} shows how the time per step (TPS) changes as more blocks are implemented on the multi-chip Loihi 2 system. The slowdown is apparent but flattens out and becomes constant for $\geq 5$ chips. This is expected as only a single activation vector has to be sent from one chip to the next at every time step (or every $N_\text{steps/block}$ steps in generation mode).

Based on the scaling shown in Figure \ref{fig:l2-chip-scaling}, we expect larger MatMul-free models to scale favorably on Loihi 2 systems, even as more layers are added.

\paragraph{Embedding and un-embedding layers} 
In our experiments, we have not implemented the embedding and un-embedding layers on the Loihi 2 chip that map between the embedding vector space $\mathbb{R}^{1024}$ and the token vocabulary of size $V=32,000$. We plan on implementing the embedding layer as a simple look-up table (LUT). The un-embedding layer is a ternary weight matrix of size $1024 \times V$ is the vocabulary size. In preliminary experiments, we mapped the un-embedding layer to 7 Loihi 2 chips, thus requiring a total of 31 Loihi 2 chips for the 370M MatMul-free language model and thus fitting onto a 32-chip system. As both embedding and un-embedding layers are single layers, they will not add significant latency to our model, thus we expect throughput to stay as reported in our experiments. The system will draw more power due to the additional layers, but we expect further performance optimizations to outweigh the power of two extra layers.
% - \textcolor{red}{NOTE: is this fine to state?}.

\paragraph{Limitations and further optimizations}
We are further optimizing latency and power of the MatMul-free LM on Loihi 2 and expect that the energy per token for the 24-chip system can approach our estimate based on a single-chip system, see Table \ref{tab:loihi-table-1-24-chip}. 
% \textcolor{red}{NOTE: do we want to add this?}

\subsubsection{Detailed NVIDIA Jetson Results}
\label{app:jetson-results-detailed}

We evaluated the inference performance and energy efficiency of several state‐of‐the‐art transformer language models running on the NVIDIA Jetson Orin Nano 8GB platform. All experiments were conducted with the MAXN power mode enabled (using Jetpack 6.2, TensorRT 10.3.0, and CUDA 12.4), and power measurements were obtained from integrated profiling tools that capture overall system consumption—including contributions from the GPU, SOC, and I/O subsystems.

Our evaluation focused on two distinct inference modes. In \textbf{prefill mode}, the model processes the input prompt in parallel. In \textbf{generate mode}, the model generates text in an autoregressive manner where tokens are produced sequentially. Due to the inherent dependencies between tokens, processing cannot be parallelized across the sequence length, resulting in lower overall throughput compared to prefill mode.

For both modes, we performed a series of inference runs (with each set averaged over 30 iterations) to ensure stable statistical estimates. The key performance metrics--throughput in tokens per second, average power consumption (in Watts), and energy per token (in Joules)--were computed by combining the timing and power measurements recorded during these runs using the \texttt{jtop} utility. 

\begin{figure}
    \centering
    \includegraphics[width=\linewidth]{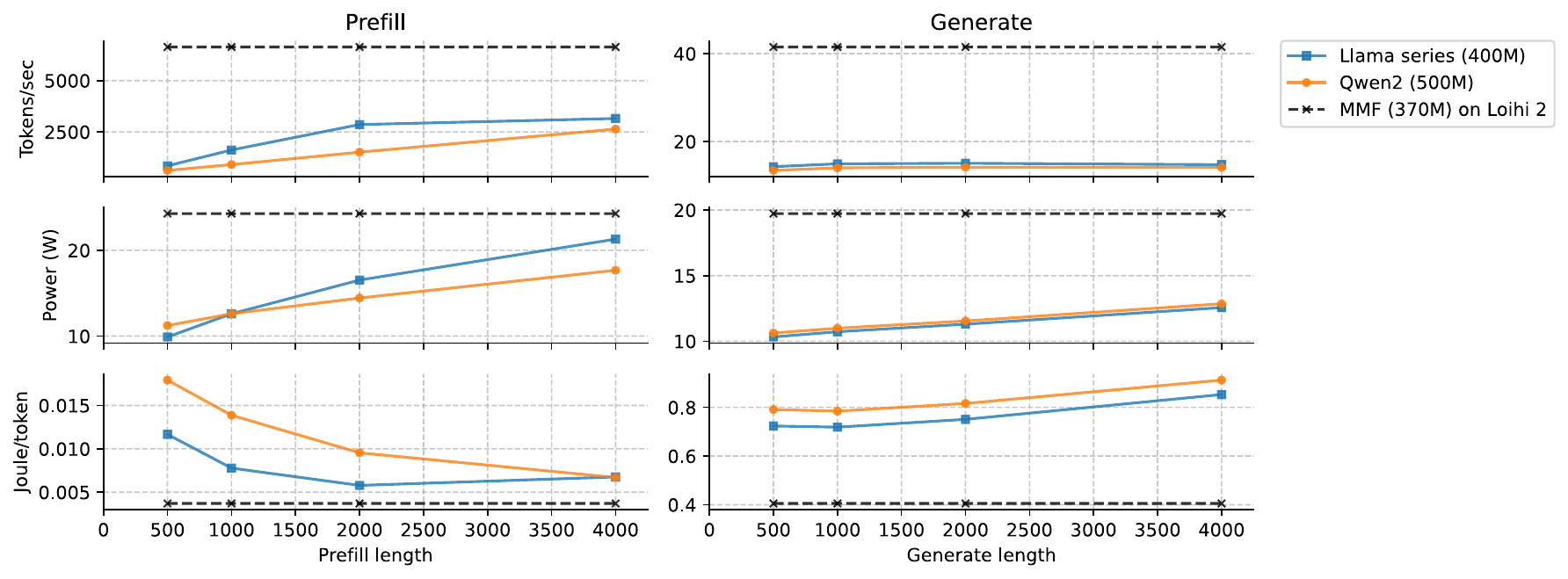}
    \caption{Hardware results for transformer-based LLMs running on the NVIDIA Jetson Orin Nano. \textit{Left}: prefill mode where text is ingested by the LLM. \textit{Right}: generate mode where text is generated in an auto-regressive loop. \textit{Top}: throughput in tokens per second. \textit{Middle}: average power in Watts. \textit{Bottom}: energy per token. All results are averaged over time for 30 inference runs. Results for the MatMul-free LLM running on Loihi 2 are based on estimates, as explained in Appendix \ref{app:l2-results-detailed}.}
    \label{fig:jetson-results-figure}
\end{figure}

In prefill mode, our measurements indicate very high throughput, reaching several thousand tokens per second. This is attributed to the parallel processing of tokens and effective pipelining of operations. In contrast, generate mode exhibits lower throughput since the sequential nature of token generation imposes latency limitations.

The dynamic power consumption is measured during active inference, reflecting contributions from all major system components. While the power draw is higher in prefill mode due to the continuous high-performance computation, generate mode incurs a more variable power profile as the system alternates between processing tasks and idling between token outputs.

Combining the timing and power data provides an estimate of the energy consumed per token. Although prefill mode can achieve high throughput, it also consumes more energy per token. In generate mode, the increased latency contributes to higher total energy per generated token, despite lower instantaneous power draw.

Collectively, these results provide a robust baseline for comparison with Loihi 2, our neuromorphic system. While transformer-based models on the Jetson Orin Nano achieve high throughput in certain configurations, they incur higher energy costs per token compared to the estimated figures for the MatMul-free LM implemented on Loihi 2. Figure~\ref{fig:jetson-results-figure} visually summarizes these metrics, displaying throughput, average power consumption, and energy per token for both prefill and generate modes.
This assessment highlights the weakness inherent in conventional transformer architectures when compared with efficient alternatives on suitable hardware.

\subsection{Model architecture on Loihi 2}
\label{app:nxkernel-graph}

\begin{figure}
    \centering
    \includegraphics[width=0.6\linewidth]{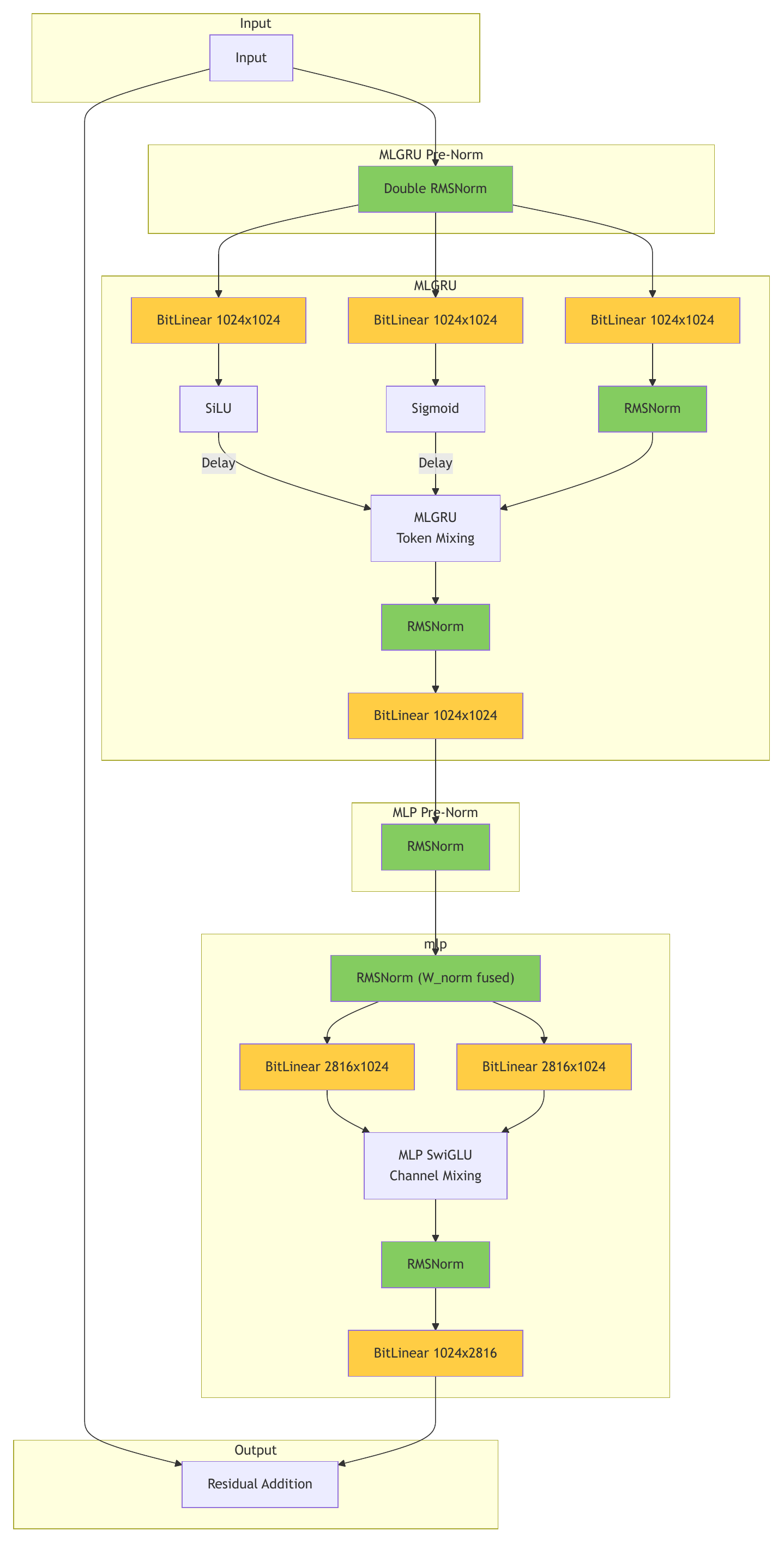}
    \includegraphics[width=0.35\linewidth]{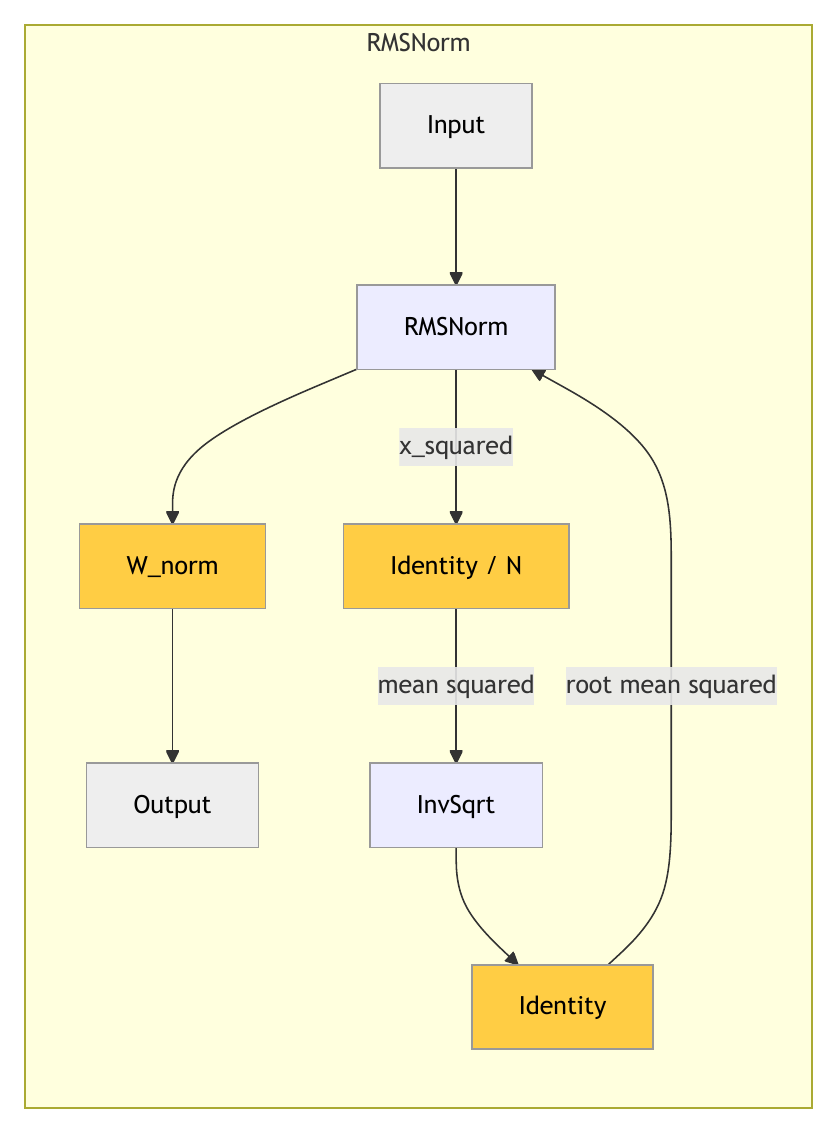}
    \caption{\textbf{Left}: Computational graph of a single MatMul-free LM layer, simplified from the actual computational graph that is mapped on the Loihi 2 chip. The RMSNorm is visualized as a single node. \textbf{Right}: Computational graph of the RMSNorm layer implemented on the Loihi 2 chip. For explanation, see main text.}
    % Neurons, synapses
    % Diagonal weights are not hidden for brevity. RMSNorm can have norm weights fused into neuron program, or separate. 
    % missing synapses are identity
    \label{fig:loihi-mmf-graph}
\end{figure}

The architecture of the MatMul-free LLM by \cite{zhu_scalable_2024} is shown in Figure \ref{fig:architecture}. As explained in Section \ref{ss:mapping-model-to-l2}, the model was mapped to the Loihi 2 architecture and the resulting computational graph is shown in Figure \ref{fig:loihi-mmf-graph}.

% RMSNorm:
% https://mermaid.ink/pdf/pako:eNp1klFPgzAUhf9Kw163OEeB2gejyXzwwZk4Ex_ELLUUWWQta4vbsuy_W6BEKHjfzjn36720nD0qEuZhL83FgWZEavC6jDlQ5eeXJEUGlP8eey9P65WQu9j7MJEtvjDBIy9K3bf_7YdVf8K43uoTuAKrfhrVp_2s99I5D3WwfhKY5G3DB4NC4z-X-m8zxpMm5gswm92aHa30jQTHjdqXRLKkyWAnMzKwMmhkaCWsyB0j5qa6cGTjqJHISlR1SyH0CGJ2aZe7OwOVkYJhIBnVl3aRcRuO29G4jcbtYNwOR22zIca4EOaFOvpeZa30ndgfxpyVUvDWgQ4A-0Al1YmTQrHWihwi6hPRYARyANQH0HBE4BBBnwiGROgQYZegOVFqyVJQdYB0m-d4giCl4dzJDQGUluKbzQ7bRGf4ujhOWych5j2kJCfMBWetjSc3dU3tuQ91TQEVuZB4Mq_LGdNcjwUYZTRNnQ77dbYlTWkCYfWTxty7_AJFwTxy?fit
% full graph:
% https://mermaid.ink/pdf/pako:eNqtVvGL2jAU_ldChbGBgolFbRkHtzmOgW6H3rEf5pDaphpWk5K0qBz735fW1mti0jm5_lBtvu99730vL6UvTsgi7PhOnLB9uA14Bp4mSwpEvt7wIN2CJDhivur_XDpfaZpnS-eXRKuL4pwzegFiGjUVdsmG55Iymz7Mn03xngTns8U3xncKLI40SAVejcsMn0g2JRQHHMA-cg_F7cZ8EEp0QTY7RiJjQgjfOiMqM06fzenQW6cb1OjHNb97Yr8xBTNyIHRzq6B7sUVXS5xdDv_XZSrp8m4qCJ0cPoLFnjxMS5uftwGlOFGNqiUgfV_RGA6bJWh0dBW97lFjjMH7Hyta_Ma5wNGHWwxe0XFT-Ll2Y7cLA2a10zFHIxnzPc8sB72E51iQKA8ScB9FJCOMmgUFrHbokePexdGuezb6l0uB6qFqF5KsCcvXCQYmOe1sg17v7vVloOFIw9EJr991CgbevbZGoUG35DVnX8U9BS8GU2o1B0-hI1UOneUUgUZlaKDhyIyfHxXQ1YKHWrDmVbWKFCtQdTbWCx-r0p6qrJQFtbKgVhbUyhopoKaMJAgmWI69wlLbAuE1LE_HqumsJ8b3_dNfdd1rWU8Zz6rVRqMkUD2pERDqUs3uW6OQNQq1RA0sZUO3DTAYKrbPlgUNbLUhk6PmlFsLt_Ubag0_HwILH1n8oDY_I1vyVuCyKmijK-wwCYSY4BgUSyAmSeJ3xm4YDvsafi_kGzbj8gOhtydRtvVheujWK1Eg5PcgD44-ZRTXy37HK69upfulvLogZAnjfqdfXlqaU5FVAA5xGMcao-paRYnjMHLd4iAtqfPnLxmRXSs?fit

\end{document}

%% file: math_commands.tex
\usepackage{amsmath,amsfonts,bm}

\def\eqref#1{equation~\ref{#1}}
\def\1{\bm{1}}

\DeclareMathAlphabet{\mathsfit}{\encodingdefault}{\sfdefault}{m}{sl}
\SetMathAlphabet{\mathsfit}{bold}{\encodingdefault}{\sfdefault}{bx}{n}

%% file: iclr2025_conference.bbl
\begin{thebibliography}{37}
\providecommand{\natexlab}[1]{#1}
\providecommand{\url}[1]{\texttt{#1}}
\expandafter\ifx\csname urlstyle\endcsname\relax
  \providecommand{\doi}[1]{doi: #1}\else
  \providecommand{\doi}{doi: \begingroup \urlstyle{rm}\Url}\fi

\bibitem[Abreu et~al.(2024)Abreu, Pedersen, Heckel, and Pierro]{abreu_q-s5_2024}
Steven Abreu, Jens~E. Pedersen, Kade~M. Heckel, and Alessandro Pierro.
\newblock Q-{S5}: {Towards} {Quantized} {State} {Space} {Models}, June 2024.
\newblock URL \url{http://arxiv.org/abs/2406.09477}.
\newblock arXiv:2406.09477 [cs].

\bibitem[Bisk et~al.(2020)Bisk, Zellers, Bras, Gao, and Choi]{bisk_piqa_2020}
Yonatan Bisk, Rowan Zellers, Ronan~Le Bras, Jianfeng Gao, and Yejin Choi.
\newblock {PIQA}: {Reasoning} about {Physical} {Commonsense} in {Natural} {Language}.
\newblock \emph{Proceedings of the AAAI Conference on Artificial Intelligence}, 34\penalty0 (05):\penalty0 7432--7439, April 2020.
\newblock ISSN 2374-3468.
\newblock \doi{10.1609/aaai.v34i05.6239}.
\newblock URL \url{https://ojs.aaai.org/index.php/AAAI/article/view/6239}.
\newblock Number: 05.

\bibitem[Chiang et~al.(2024)Chiang, Chang, Frumkin, Wu, and Marculescu]{chiang_quamba_2024}
Hung-Yueh Chiang, Chi-Chih Chang, Natalia Frumkin, Kai-Chiang Wu, and Diana Marculescu.
\newblock Quamba: {A} {Post}-{Training} {Quantization} {Recipe} for {Selective} {State} {Space} {Models}, October 2024.
\newblock URL \url{http://arxiv.org/abs/2410.13229}.
\newblock arXiv:2410.13229.

\bibitem[Cho et~al.(2014)Cho, van Merriënboer, Gulcehre, Bahdanau, Bougares, Schwenk, and Bengio]{cho_learning_2014}
Kyunghyun Cho, Bart van Merriënboer, Caglar Gulcehre, Dzmitry Bahdanau, Fethi Bougares, Holger Schwenk, and Yoshua Bengio.
\newblock Learning {Phrase} {Representations} using {RNN} {Encoder}–{Decoder} for {Statistical} {Machine} {Translation}.
\newblock In Alessandro Moschitti, Bo~Pang, and Walter Daelemans (eds.), \emph{Proceedings of the 2014 {Conference} on {Empirical} {Methods} in {Natural} {Language} {Processing} ({EMNLP})}, pp.\  1724--1734, Doha, Qatar, October 2014. Association for Computational Linguistics.
\newblock \doi{10.3115/v1/D14-1179}.
\newblock URL \url{https://aclanthology.org/D14-1179}.

\bibitem[Clark et~al.(2018)Clark, Cowhey, Etzioni, Khot, Sabharwal, Schoenick, and Tafjord]{clark_think_2018}
Peter Clark, Isaac Cowhey, Oren Etzioni, Tushar Khot, Ashish Sabharwal, Carissa Schoenick, and Oyvind Tafjord.
\newblock Think you have {Solved} {Question} {Answering}? {Try} {ARC}, the {AI2} {Reasoning} {Challenge}, March 2018.
\newblock URL \url{http://arxiv.org/abs/1803.05457}.
\newblock arXiv:1803.05457.

\bibitem[Courbariaux et~al.(2016)Courbariaux, Hubara, Soudry, El-Yaniv, and Bengio]{courbariaux_binarized_2016}
Matthieu Courbariaux, Itay Hubara, Daniel Soudry, Ran El-Yaniv, and Yoshua Bengio.
\newblock Binarized {Neural} {Networks}: {Training} {Deep} {Neural} {Networks} with {Weights} and {Activations} {Constrained} to +1 or -1, March 2016.
\newblock URL \url{http://arxiv.org/abs/1602.02830}.
\newblock arXiv:1602.02830 [cs].

\bibitem[Dauphin et~al.(2017)Dauphin, Fan, Auli, and Grangier]{dauphin_language_2017}
Yann~N. Dauphin, Angela Fan, Michael Auli, and David Grangier.
\newblock Language modeling with gated convolutional networks.
\newblock In \emph{Proceedings of the 34th {International} {Conference} on {Machine} {Learning} - {Volume} 70}, {ICML}'17, pp.\  933--941, Sydney, NSW, Australia, August 2017. JMLR.org.

\bibitem[Davies et~al.(2021)Davies, Wild, Orchard, Sandamirskaya, Guerra, Joshi, Plank, and Risbud]{davies_advancing_2021}
Mike Davies, Andreas Wild, Garrick Orchard, Yulia Sandamirskaya, Gabriel A.~Fonseca Guerra, Prasad Joshi, Philipp Plank, and Sumedh~R. Risbud.
\newblock Advancing {Neuromorphic} {Computing} {With} {Loihi}: {A} {Survey} of {Results} and {Outlook}.
\newblock \emph{Proceedings of the IEEE}, 109\penalty0 (5):\penalty0 911--934, May 2021.
\newblock \doi{10.1109/jproc.2021.3067593}.
\newblock Publisher: Institute of Electrical and Electronics Engineers (IEEE).

\bibitem[DeepSeek-AI et~al.(2025)DeepSeek-AI, Guo, Yang, Zhang, Song, Zhang, Xu, Zhu, Ma, Wang, Bi, Zhang, Yu, Wu, Wu, Gou, Shao, Li, Gao, Liu, Xue, Wang, Wu, Feng, Lu, Zhao, Deng, Zhang, Ruan, Dai, Chen, Ji, Li, Lin, Dai, Luo, Hao, Chen, Li, Zhang, Bao, Xu, Wang, Ding, Xin, Gao, Qu, Li, Guo, Li, Wang, Chen, Yuan, Qiu, Li, Cai, Ni, Liang, Chen, Dong, Hu, Gao, Guan, Huang, Yu, Wang, Zhang, Zhao, Wang, Zhang, Xu, Xia, Zhang, Zhang, Tang, Li, Wang, Li, Tian, Huang, Zhang, Wang, Chen, Du, Ge, Zhang, Pan, Wang, Chen, Jin, Chen, Lu, Zhou, Chen, Ye, Wang, Yu, Zhou, Pan, Li, Zhou, Wu, Ye, Yun, Pei, Sun, Wang, Zeng, Zhao, Liu, Liang, Gao, Yu, Zhang, Xiao, An, Liu, Wang, Chen, Nie, Cheng, Liu, Xie, Liu, Yang, Li, Su, Lin, Li, Jin, Shen, Chen, Sun, Wang, Song, Zhou, Wang, Shan, Li, Wang, Wei, Zhang, Xu, Li, Zhao, Sun, Wang, Yu, Zhang, Shi, Xiong, He, Piao, Wang, Tan, Ma, Liu, Guo, Ou, Wang, Gong, Zou, He, Xiong, Luo, You, Liu, Zhou, Zhu, Xu, Huang, Li, Zheng, Zhu, Ma, Tang, Zha, Yan, Ren, Ren, Sha, Fu, Xu, Xie, Zhang,
  Hao, Ma, Yan, Wu, Gu, Zhu, Liu, Li, Xie, Song, Pan, Huang, Xu, Zhang, and Zhang]{deepseek-ai_deepseek-r1_2025}
DeepSeek-AI, Daya Guo, Dejian Yang, Haowei Zhang, Junxiao Song, Ruoyu Zhang, Runxin Xu, Qihao Zhu, Shirong Ma, Peiyi Wang, Xiao Bi, Xiaokang Zhang, Xingkai Yu, Yu~Wu, Z.~F. Wu, Zhibin Gou, Zhihong Shao, Zhuoshu Li, Ziyi Gao, Aixin Liu, Bing Xue, Bingxuan Wang, Bochao Wu, Bei Feng, Chengda Lu, Chenggang Zhao, Chengqi Deng, Chenyu Zhang, Chong Ruan, Damai Dai, Deli Chen, Dongjie Ji, Erhang Li, Fangyun Lin, Fucong Dai, Fuli Luo, Guangbo Hao, Guanting Chen, Guowei Li, H.~Zhang, Han Bao, Hanwei Xu, Haocheng Wang, Honghui Ding, Huajian Xin, Huazuo Gao, Hui Qu, Hui Li, Jianzhong Guo, Jiashi Li, Jiawei Wang, Jingchang Chen, Jingyang Yuan, Junjie Qiu, Junlong Li, J.~L. Cai, Jiaqi Ni, Jian Liang, Jin Chen, Kai Dong, Kai Hu, Kaige Gao, Kang Guan, Kexin Huang, Kuai Yu, Lean Wang, Lecong Zhang, Liang Zhao, Litong Wang, Liyue Zhang, Lei Xu, Leyi Xia, Mingchuan Zhang, Minghua Zhang, Minghui Tang, Meng Li, Miaojun Wang, Mingming Li, Ning Tian, Panpan Huang, Peng Zhang, Qiancheng Wang, Qinyu Chen, Qiushi Du, Ruiqi Ge, Ruisong
  Zhang, Ruizhe Pan, Runji Wang, R.~J. Chen, R.~L. Jin, Ruyi Chen, Shanghao Lu, Shangyan Zhou, Shanhuang Chen, Shengfeng Ye, Shiyu Wang, Shuiping Yu, Shunfeng Zhou, Shuting Pan, S.~S. Li, Shuang Zhou, Shaoqing Wu, Shengfeng Ye, Tao Yun, Tian Pei, Tianyu Sun, T.~Wang, Wangding Zeng, Wanjia Zhao, Wen Liu, Wenfeng Liang, Wenjun Gao, Wenqin Yu, Wentao Zhang, W.~L. Xiao, Wei An, Xiaodong Liu, Xiaohan Wang, Xiaokang Chen, Xiaotao Nie, Xin Cheng, Xin Liu, Xin Xie, Xingchao Liu, Xinyu Yang, Xinyuan Li, Xuecheng Su, Xuheng Lin, X.~Q. Li, Xiangyue Jin, Xiaojin Shen, Xiaosha Chen, Xiaowen Sun, Xiaoxiang Wang, Xinnan Song, Xinyi Zhou, Xianzu Wang, Xinxia Shan, Y.~K. Li, Y.~Q. Wang, Y.~X. Wei, Yang Zhang, Yanhong Xu, Yao Li, Yao Zhao, Yaofeng Sun, Yaohui Wang, Yi~Yu, Yichao Zhang, Yifan Shi, Yiliang Xiong, Ying He, Yishi Piao, Yisong Wang, Yixuan Tan, Yiyang Ma, Yiyuan Liu, Yongqiang Guo, Yuan Ou, Yuduan Wang, Yue Gong, Yuheng Zou, Yujia He, Yunfan Xiong, Yuxiang Luo, Yuxiang You, Yuxuan Liu, Yuyang Zhou, Y.~X. Zhu,
  Yanhong Xu, Yanping Huang, Yaohui Li, Yi~Zheng, Yuchen Zhu, Yunxian Ma, Ying Tang, Yukun Zha, Yuting Yan, Z.~Z. Ren, Zehui Ren, Zhangli Sha, Zhe Fu, Zhean Xu, Zhenda Xie, Zhengyan Zhang, Zhewen Hao, Zhicheng Ma, Zhigang Yan, Zhiyu Wu, Zihui Gu, Zijia Zhu, Zijun Liu, Zilin Li, Ziwei Xie, Ziyang Song, Zizheng Pan, Zhen Huang, Zhipeng Xu, Zhongyu Zhang, and Zhen Zhang.
\newblock {DeepSeek}-{R1}: {Incentivizing} {Reasoning} {Capability} in {LLMs} via {Reinforcement} {Learning}, January 2025.
\newblock URL \url{http://arxiv.org/abs/2501.12948}.
\newblock arXiv:2501.12948 [cs].

\bibitem[Dettmers et~al.(2022)Dettmers, Lewis, Belkada, and Zettlemoyer]{dettmers_llmint8_2022}
Tim Dettmers, Mike Lewis, Younes Belkada, and Luke Zettlemoyer.
\newblock {LLM}.int8(): 8-bit {Matrix} {Multiplication} for {Transformers} at {Scale}, November 2022.
\newblock URL \url{http://arxiv.org/abs/2208.07339}.
\newblock arXiv:2208.07339 [cs].

\bibitem[Frantar et~al.(2023)Frantar, Ashkboos, Hoefler, and Alistarh]{frantar_gptq_2023}
Elias Frantar, Saleh Ashkboos, Torsten Hoefler, and Dan Alistarh.
\newblock {GPTQ}: {Accurate} {Post}-{Training} {Quantization} for {Generative} {Pre}-trained {Transformers}, March 2023.
\newblock URL \url{http://arxiv.org/abs/2210.17323}.
\newblock arXiv:2210.17323 [cs].

\bibitem[Gu \& Dao(2023)Gu and Dao]{gu2023mamba}
Albert Gu and Tri Dao.
\newblock Mamba: Linear-time sequence modeling with selective state spaces.
\newblock \emph{arXiv preprint arXiv:2312.00752}, 2023.

\bibitem[Gupta et~al.(2022)Gupta, Gu, and Berant]{gupta2022diagonal}
Ankit Gupta, Albert Gu, and Jonathan Berant.
\newblock Diagonal state spaces are as effective as structured state spaces.
\newblock \emph{Advances in Neural Information Processing Systems}, 35:\penalty0 22982--22994, 2022.

\bibitem[Kosson \& Jaggi(2023)Kosson and Jaggi]{kosson_multiplication-free_2023}
Atli Kosson and Martin Jaggi.
\newblock Multiplication-{Free} {Transformer} {Training} via {Piecewise} {Affine} {Operations}, October 2023.
\newblock URL \url{http://arxiv.org/abs/2305.17190}.
\newblock arXiv:2305.17190 [cs].

\bibitem[Ma et~al.(2024)Ma, Wang, Ma, Wang, Wang, Huang, Dong, Wang, Xue, and Wei]{ma_era_2024}
Shuming Ma, Hongyu Wang, Lingxiao Ma, Lei Wang, Wenhui Wang, Shaohan Huang, Li~Dong, Ruiping Wang, Jilong Xue, and Furu Wei.
\newblock The {Era} of 1-bit {LLMs}: {All} {Large} {Language} {Models} are in 1.58 {Bits}, February 2024.
\newblock URL \url{http://arxiv.org/abs/2402.17764}.
\newblock arXiv:2402.17764 [cs].

\bibitem[Mihaylov et~al.(2018)Mihaylov, Clark, Khot, and Sabharwal]{mihaylov_can_2018}
Todor Mihaylov, Peter Clark, Tushar Khot, and Ashish Sabharwal.
\newblock Can a {Suit} of {Armor} {Conduct} {Electricity}? {A} {New} {Dataset} for {Open} {Book} {Question} {Answering}.
\newblock In \emph{{EMNLP}}, 2018.

\bibitem[Montebovi(2024)]{alireo2024}
Michele Montebovi.
\newblock Alireo-400m: A lightweight italian language model, 2024.
\newblock URL \url{https://huggingface.co/DeepMount00/Alireo-400m-instruct-v0.1}.

\bibitem[Niu et~al.(2021)Niu, Guan, Wang, Agrawal, and Ren]{niu_dnnfusion_2021}
Wei Niu, Jiexiong Guan, Yanzhi Wang, Gagan Agrawal, and Bin Ren.
\newblock Dnnfusion: accelerating deep neural networks execution with advanced operator fusion.
\newblock In \emph{Proceedings of the 42nd ACM SIGPLAN International Conference on Programming Language Design and Implementation}, PLDI 2021, pp.\  883–898, New York, NY, USA, 2021. Association for Computing Machinery.
\newblock ISBN 9781450383912.
\newblock \doi{10.1145/3453483.3454083}.
\newblock URL \url{https://doi.org/10.1145/3453483.3454083}.

\bibitem[Orchard et~al.(2021{\natexlab{a}})Orchard, Frady, Rubin, Sanborn, Shrestha, Sommer, and Davies]{orchard2021efficient}
Garrick Orchard, E~Paxon Frady, Daniel Ben~Dayan Rubin, Sophia Sanborn, Sumit~Bam Shrestha, Friedrich~T Sommer, and Mike Davies.
\newblock Efficient neuromorphic signal processing with loihi 2.
\newblock In \emph{2021 IEEE Workshop on Signal Processing Systems (SiPS)}, pp.\  254--259. IEEE, 2021{\natexlab{a}}.

\bibitem[Orchard et~al.(2021{\natexlab{b}})Orchard, Frady, Rubin, Sanborn, Shrestha, Sommer, and Davies]{orchard_efficient_2021}
Garrick Orchard, E.~Paxon Frady, Daniel Ben~Dayan Rubin, Sophia Sanborn, Sumit~Bam Shrestha, Friedrich~T. Sommer, and Mike Davies.
\newblock Efficient {Neuromorphic} {Signal} {Processing} with {Loihi} 2.
\newblock In \emph{2021 {IEEE} {Workshop} on {Signal} {Processing} {Systems} ({SiPS})}. IEEE, October 2021{\natexlab{b}}.
\newblock \doi{10.1109/sips52927.2021.00053}.

\bibitem[Pierro \& Abreu(2024)Pierro and Abreu]{pierro_mamba-ptq_2024}
Alessandro Pierro and Steven Abreu.
\newblock Mamba-{PTQ}: {Outlier} {Channels} in {Recurrent} {Large} {Language} {Models}, July 2024.
\newblock URL \url{http://arxiv.org/abs/2407.12397}.
\newblock arXiv:2407.12397 [cs].

\bibitem[Qin et~al.(2023)Qin, Yang, and Zhong]{qin_hierarchically_2023}
Zhen Qin, Songlin Yang, and Yiran Zhong.
\newblock Hierarchically {Gated} {Recurrent} {Neural} {Network} for {Sequence} {Modeling}.
\newblock \emph{Advances in Neural Information Processing Systems}, 36:\penalty0 33202--33221, December 2023.

\bibitem[Qin et~al.(2024)Qin, Yang, Sun, Shen, Li, Sun, and Zhong]{qin_hgrn2_2024}
Zhen Qin, Songlin Yang, Weixuan Sun, Xuyang Shen, Dong Li, Weigao Sun, and Yiran Zhong.
\newblock {HGRN2}: {Gated} {Linear} {RNNs} with {State} {Expansion}, April 2024.
\newblock URL \url{http://arxiv.org/abs/2404.07904}.
\newblock arXiv:2404.07904 [cs].

\bibitem[{Qwen Team}(2024)]{qwen2.5}
{Qwen Team}.
\newblock Qwen2.5: A party of foundation models, September 2024.
\newblock URL \url{https://qwenlm.github.io/blog/qwen2.5/}.

\bibitem[Sakaguchi et~al.(2021)Sakaguchi, Bras, Bhagavatula, and Choi]{sakaguchi_winogrande_2021}
Keisuke Sakaguchi, Ronan~Le Bras, Chandra Bhagavatula, and Yejin Choi.
\newblock {WinoGrande}: an adversarial winograd schema challenge at scale.
\newblock \emph{Commun. ACM}, 64\penalty0 (9):\penalty0 99--106, August 2021.
\newblock ISSN 0001-0782.
\newblock \doi{10.1145/3474381}.
\newblock URL \url{https://dl.acm.org/doi/10.1145/3474381}.

\bibitem[Shrestha et~al.(2024)Shrestha, Timcheck, Frady, Campos-Macias, and Davies]{shrestha_efficient_2024}
Sumit~Bam Shrestha, Jonathan Timcheck, Paxon Frady, Leobardo Campos-Macias, and Mike Davies.
\newblock Efficient {Video} and {Audio} {Processing} with {Loihi} 2.
\newblock In \emph{{ICASSP} 2024 - 2024 {IEEE} {International} {Conference} on {Acoustics}, {Speech} and {Signal} {Processing} ({ICASSP})}, pp.\  13481--13485, April 2024.
\newblock \doi{10.1109/ICASSP48485.2024.10448003}.
\newblock URL \url{https://ieeexplore.ieee.org/abstract/document/10448003}.
\newblock ISSN: 2379-190X.

\bibitem[Vaswani et~al.(2017)Vaswani, Shazeer, Parmar, Uszkoreit, Jones, Gomez, Kaiser, and Polosukhin]{vaswani_attention_2017}
Ashish Vaswani, Noam Shazeer, Niki Parmar, Jakob Uszkoreit, Llion Jones, Aidan~N. Gomez, Lukasz Kaiser, and Illia Polosukhin.
\newblock Attention {Is} {All} {You} {Need}.
\newblock June 2017.
\newblock URL \url{https://nlp.seas.harvard.edu/2018/04/03/attention.html}.
\newblock \_eprint: 1706.03762.

\bibitem[Waeijen et~al.(2021)Waeijen, Sioutas, Peemen, Lindwer, and Corporaal]{waeijen_convfusion_2021}
Luc Waeijen, Savvas Sioutas, Maurice Peemen, Menno Lindwer, and Henk Corporaal.
\newblock Convfusion: A model for layer fusion in convolutional neural networks.
\newblock \emph{IEEE Access}, 9:\penalty0 168245--168267, 2021.
\newblock \doi{10.1109/ACCESS.2021.3134930}.

\bibitem[Xiao et~al.(2024)Xiao, Lin, Seznec, Wu, Demouth, and Han]{xiao_smoothquant_2024}
Guangxuan Xiao, Ji~Lin, Mickael Seznec, Hao Wu, Julien Demouth, and Song Han.
\newblock {SmoothQuant}: {Accurate} and {Efficient} {Post}-{Training} {Quantization} for {Large} {Language} {Models}, March 2024.
\newblock URL \url{http://arxiv.org/abs/2211.10438}.
\newblock arXiv:2211.10438 [cs].

\bibitem[Yang et~al.(2024)Yang, Yang, Hui, Zheng, Yu, Zhou, Li, Li, Liu, Huang, Dong, Wei, Lin, Tang, Wang, Yang, Tu, Zhang, Ma, Xu, Zhou, Bai, He, Lin, Dang, Lu, Chen, Yang, Li, Xue, Ni, Zhang, Wang, Peng, Men, Gao, Lin, Wang, Bai, Tan, Zhu, Li, Liu, Ge, Deng, Zhou, Ren, Zhang, Wei, Ren, Fan, Yao, Zhang, Wan, Chu, Liu, Cui, Zhang, and Fan]{qwen2}
An~Yang, Baosong Yang, Binyuan Hui, Bo~Zheng, Bowen Yu, Chang Zhou, Chengpeng Li, Chengyuan Li, Dayiheng Liu, Fei Huang, Guanting Dong, Haoran Wei, Huan Lin, Jialong Tang, Jialin Wang, Jian Yang, Jianhong Tu, Jianwei Zhang, Jianxin Ma, Jin Xu, Jingren Zhou, Jinze Bai, Jinzheng He, Junyang Lin, Kai Dang, Keming Lu, Keqin Chen, Kexin Yang, Mei Li, Mingfeng Xue, Na~Ni, Pei Zhang, Peng Wang, Ru~Peng, Rui Men, Ruize Gao, Runji Lin, Shijie Wang, Shuai Bai, Sinan Tan, Tianhang Zhu, Tianhao Li, Tianyu Liu, Wenbin Ge, Xiaodong Deng, Xiaohuan Zhou, Xingzhang Ren, Xinyu Zhang, Xipin Wei, Xuancheng Ren, Yang Fan, Yang Yao, Yichang Zhang, Yu~Wan, Yunfei Chu, Yuqiong Liu, Zeyu Cui, Zhenru Zhang, and Zhihao Fan.
\newblock Qwen2 technical report.
\newblock \emph{arXiv preprint arXiv:2407.10671}, 2024.

\bibitem[You et~al.(2024)You, Guo, Fu, Zhou, Shi, Zhang, Kundu, Yazdanbakhsh, and Lin]{you_shiftaddllm_2024}
Haoran You, Yipin Guo, Yichao Fu, Wei Zhou, Huihong Shi, Xiaofan Zhang, Souvik Kundu, Amir Yazdanbakhsh, and Yingyan~Celine Lin.
\newblock {ShiftAddLLM}: {Accelerating} {Pretrained} {LLMs} via {Post}-{Training} {Multiplication}-{Less} {Reparameterization}, July 2024.
\newblock URL \url{http://arxiv.org/abs/2406.05981}.
\newblock arXiv:2406.05981 [cs] version: 3.

\bibitem[Yu et~al.(2022)Yu, Luo, Zhou, Si, Zhou, Wang, Feng, and Yan]{Yu_2022_CVPR}
Weihao Yu, Mi~Luo, Pan Zhou, Chenyang Si, Yichen Zhou, Xinchao Wang, Jiashi Feng, and Shuicheng Yan.
\newblock Metaformer is actually what you need for vision.
\newblock In \emph{Proceedings of the IEEE/CVF Conference on Computer Vision and Pattern Recognition (CVPR)}, pp.\  10819--10829, June 2022.

\bibitem[Zellers et~al.(2019)Zellers, Holtzman, Bisk, Farhadi, and Choi]{zellers_hellaswag_2019}
Rowan Zellers, Ari Holtzman, Yonatan Bisk, Ali Farhadi, and Yejin Choi.
\newblock {HellaSwag}: {Can} a {Machine} {Really} {Finish} {Your} {Sentence}?
\newblock In Anna Korhonen, David Traum, and Lluís Màrquez (eds.), \emph{Proceedings of the 57th {Annual} {Meeting} of the {Association} for {Computational} {Linguistics}}, pp.\  4791--4800, Florence, Italy, July 2019. Association for Computational Linguistics.
\newblock \doi{10.18653/v1/P19-1472}.
\newblock URL \url{https://aclanthology.org/P19-1472}.

\bibitem[Zhang \& Sennrich(2019)Zhang and Sennrich]{zhang_root_2019}
Biao Zhang and Rico Sennrich.
\newblock Root {Mean} {Square} {Layer} {Normalization}.
\newblock In \emph{Advances in {Neural} {Information} {Processing} {Systems}}, volume~32. Curran Associates, Inc., 2019.

\bibitem[Zhang et~al.(2023)Zhang, Garg, Cao, Lew, Ghorbani, Zhang, and Firat]{zhang_binarized_2023}
Yichi Zhang, Ankush Garg, Yuan Cao, Lukasz Lew, Behrooz Ghorbani, Zhiru Zhang, and Orhan Firat.
\newblock Binarized {Neural} {Machine} {Translation}, February 2023.
\newblock URL \url{http://arxiv.org/abs/2302.04907}.
\newblock arXiv:2302.04907 [cs].

\bibitem[Zhu et~al.(2023)Zhu, Zhao, Li, and Eshraghian]{zhu2023spikegpt}
Rui-Jie Zhu, Qihang Zhao, Guoqi Li, and Jason~K Eshraghian.
\newblock Spikegpt: Generative pre-trained language model with spiking neural networks.
\newblock \emph{arXiv preprint arXiv:2302.13939}, 2023.

\bibitem[Zhu et~al.(2024)Zhu, Zhang, Sifferman, Sheaves, Wang, Richmond, Zhou, and Eshraghian]{zhu_scalable_2024}
Rui-Jie Zhu, Yu~Zhang, Ethan Sifferman, Tyler Sheaves, Yiqiao Wang, Dustin Richmond, Peng Zhou, and Jason~K. Eshraghian.
\newblock Scalable {MatMul}-free {Language} {Modeling}, 2024.
\newblock URL \url{http://arxiv.org/abs/2406.02528}.
\newblock arXiv:2406.02528 [cs].

\end{thebibliography}
